\documentclass[journal]{IEEEtran}
\ifCLASSINFOpdf
\else
\fi

\usepackage{graphicx}
\usepackage{amsmath}
\usepackage{array}
\usepackage{url}
\usepackage{multirow}
\usepackage{subcaption,booktabs}
\renewcommand{\vec}[1]{\mathbf{#1}}
\usepackage{float}
\usepackage{placeins}
\usepackage{xcolor}

\begin{document}
\title{A Distance Map Regularized CNN for Cardiac Cine MR Image Segmentation}

\author{Shusil~Dangi,
        Cristian~A.~Linte,~\IEEEmembership{Senior Member,~IEEE,}
        and~Ziv~Yaniv,~\IEEEmembership{Senior~Member,~IEEE}
% <-this % stops a space
% \thanks{Manuscript received January, 2019; revised xx, 2019.}%
\thanks{This work has been submitted to the IEEE for possible publication. Copyright may be transferred without notice, after which this version may no longer be accessible.}%
\thanks{Shusil Dangi is with the Center for Imaging Science, Rochester Institute of Technology, Rochester,
NY USA. E-mail: sxd7257@rit.edu.}% <-this % stops a space
\thanks{Cristian A. Linte is with the Biomedical Engineering and Center for Imaging Science, Rochester Institute of Technology, Rochester NY USA. Email: calbme@rit.edu}
\thanks{Ziv Yaniv is with the National Institute of Allergy and Infectious Diseases, Bethesda MD USA and MSC LLC., Rockville MD USA. E-mail: zivyaniv@nih.gov}}
\maketitle

\begin{abstract}
Cardiac image segmentation is a critical process for generating personalized models of the heart and for quantifying cardiac performance parameters. Several convolutional neural network (CNN) architectures have been proposed to segment the heart chambers from cardiac cine MR images. Here we propose a multi-task learning (MTL)-based regularization framework for cardiac MR image segmentation. The network is trained to perform the main task of semantic segmentation, along with a simultaneous, auxiliary task of pixel-wise distance map regression. The proposed distance map regularizer is a decoder network added to the bottleneck layer of an existing CNN architecture, facilitating the network to learn robust global features. The regularizer block is removed after training, so that the original number of network parameters does not change. We show that the proposed regularization method improves both binary and multi-class segmentation performance over the corresponding state-of-the-art CNN architectures on two publicly available cardiac cine MRI datasets, obtaining average dice coefficient of 0.84$\pm$0.03 and 0.91$\pm$0.04, respectively. Furthermore, we also demonstrate improved generalization performance of the distance map regularized network on cross-dataset segmentation, showing as much as 42\% improvement in myocardium Dice coefficient from 0.56$\pm$0.28 to 0.80$\pm$0.14.
\end{abstract}


% \begin{IEEEkeywords}
% Convolutional Neural Network, Distance Transform, Regularization, Medical Image Segmentation, Multi-Task Learning, Generalization, Transfer Learning.
% \end{IEEEkeywords}

% \fi
\IEEEpeerreviewmaketitle

\section{Introduction}
\IEEEPARstart{M}{agnetic} Resonance Imaging (MRI) is the standard-of-care imaging modality for non-invasive cardiac diagnosis, due to its high contrast sensitivity to soft tissue, good image quality, and lack of exposure to ionizing radiation. Cine cardiac MRI enables the acquisition of high resolution two-dimensional (2D) anatomical images of the heart throughout the cardiac cycle, capturing the full cardiac dynamics via multiple 2D + time short axis acquisitions spanning the whole heart. 

\begin{figure}[t]
\centering
\includegraphics[width=0.47\textwidth]{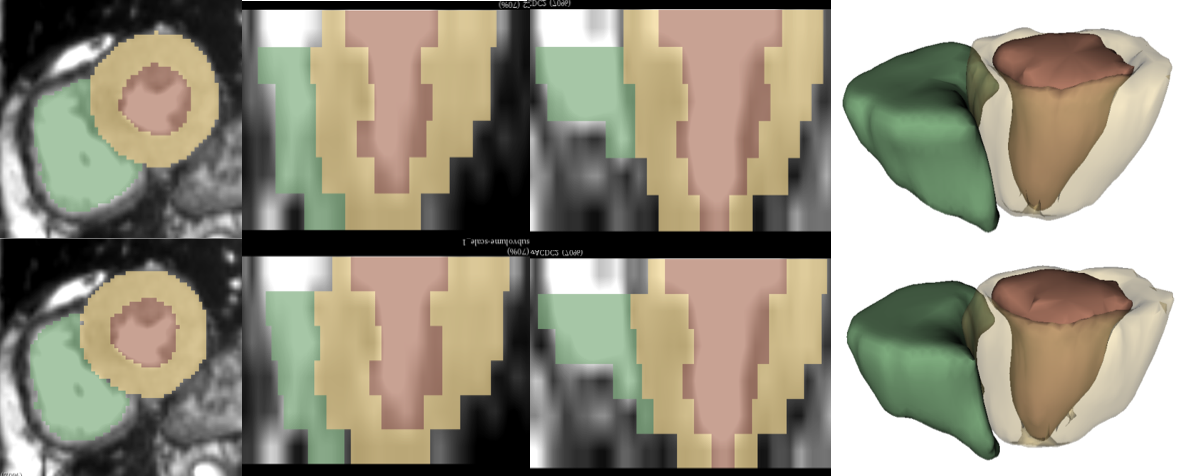}
\caption{Segmentation results for LV blood-pool, LV myocardium, and RV blood-pool. First column shows the short-axis view, second and third columns show orthogonal long-axis views, and the fourth column shows generated three dimensional models. Reference (top row) and segmentation obtained from the DMR-UNet model (bottom row).}
\label{fig:SegmentationResults}
\vspace{-4mm}
\end{figure}

Segmentation of the heart structures from these images enables measurement of important cardiac diagnostic indices such as myocardial mass and thickness, left/right ventricle (LV/RV) volumes and ejection fraction. Furthermore, high-quality personalized heart models can be generated for cardiac morphology assessment, treatment planning, as well as, precise localization of pathologies during an image-guided intervention. Manual delineation is the standard cardiac image segmentation approach, which is not only time consuming, but also susceptible to high inter- and intra-observer variability. Hence, there is a critical need for semi-/fully-automatic methods for cardiac cine MRI segmentation. However, the MR imaging artifacts such as bias fields, respiratory motion, and intensity inhomogeneity and fuzziness, render the segmentation of heart structures challenging. {\bf Fig. \ref{fig:SegmentationResults}} shows a reference segmentation and the results of our automatic segmentation method.

A comprehensive review of cardiac MR segmentation techniques can be found in \cite{Peng:2016,Petitjean2011}. These techniques can be classified based on the amount of prior knowledge used during segmentation. First, the {\it no-prior} based methods rely solely on the image content to segment the heart structures based on intensity thresholds, and edge- and/or region-information. Hence these methods are often ineffective for the segmentation of ill-defined boundary regions. Second, the deformable models such as active contours and level-set methods incorporate {\it weak-prior} information regarding the smoothness of the segmented boundaries; similarly, graph theoretical models assume connectivity between the neighboring pixels providing piece-wise smooth segmentation results. Third, the Active shape and appearance models and Atlas-based methods impose very {\it strong-prior} information regarding the geometry of the heart structures and sometimes are too restricted by the training set. These {\it weak-/strong-prior} based methods may overcome segmentation challenges in ill-defined boundary regions but, nevertheless, at a high computational cost. Lastly, {\it Machine Learning} based methods aim to predict the probability of each pixel in the image belonging to the foreground/background class based on either patch-wise or image-wise training. These methods are able to produce fast and accurate segmentation, provided the training set captures the population variability.

In the context of deep learning, Long {\it et al.} \cite{Long:2015} proposed the first fully convolutional network (FCN) for semantic image segmentation, exploiting the capability of Convolutional Neural Networks (CNNs) \cite{LeCun:1998,Goodfellow:2016,LeCun:2015} to learn task-specific hierarchical features in an end-to-end manner. However, their initial adoption in the medical domain was challenging, due to the limited availability of medical imaging data and associated costly manual annotation. These challenges were later circumvented by patch-based training, data augmentation, and transfer learning techniques \cite{Shen:2017,Litjens:2017}. 

Specifically, in the context of cardiac image segmentation, Tran \cite{Tran:2016} adapted a FCN architecture for segmentation of various cardiac structures from short-axis MR images. Similarly, Poudel {\it et al.} \cite{Poudel:2016} proposed a recurrent FCN architecture to leverage inter-slice spatial dependencies between the 2D cine MR slices. Avendi {\it et al.} \cite{Avendi:2016} reported improved accuracy and robustness of the LV segmentation by using the output of a FCN to initialize a deformable model. Further, Oktay {\it et al.} \cite{Oktay:2018} pre-trained an auto-encoder network on ground-truth segmentations and imposed anatomical constraints into a CNN network by adding {\it $l_{2}$-}loss between the auto-encoder representation of the output and the corresponding ground-truth segmentation. Several modifications to the FCN architecture and various post-processing schemes have been proposed to improve the semantic segmentation results as summarized in \cite{Garcia:2017}.

To improve the generalization performance of neural networks, various regularization techniques have been proposed. These include parameter norm penalty (e.g. weight decay \cite{Krogh:1991}), noise injection \cite{Vincent:2008}, dropout \cite{Srivastava:2014}, batch normalization \cite{Ioffe:2015}, adversarial training \cite{Goodfellow:2015}, and multi-task learning (MTL) \cite{Caruana:1997}. In this paper we focus on MTL-based network regularization. When a network is trained on multiple related tasks, the inductive bias provided by the auxiliary tasks causes the model to prefer a hypothesis that explains more than one task. This helps the network ignore task-specific noise and hence focus on learning features relevant to multiple tasks, improving the generalization performance \cite{Caruana:1997}. Furthermore, MTL reduces the Rademacher complexity \cite{Bartlett:2003} of the model (i.e. its ability to fit random noise), hence reducing the risk of overfitting. An overview of MTL applied to deep neural networks can be found in \cite{Ruder:2017}.

MTL has been widely employed in computer vision problems due to the similarity between various tasks being performed. A FCN architecture with a common encoder and task specific decoders was proposed in \cite{Teichmann:2018} to perform joint classification, detection, and semantic segmentation, targeting real-time applications such as autonomous driving. A similar single-encoder-multiple-decoder architecture described in \cite{Uhrig:2016} performs semantic segmentation, depth regression, and instance segmentation, simultaneously. The architecture was further expanded by \cite{Kendall:2017MTL} to automatically learn the weights for each task based on its uncertainty, obtaining state-of-the-art results. 

In the context of medical image analysis, Moeskops {\it et al.} \cite{Moeskops:2016} demonstrated the use of MTL for joint segmentation of six tissue types from brain MRI, the pectoral muscle from breast MRI, and the coronary arteries from cardiac Computed Tomography Angiography (CTA) images, with performance equivalent to networks trained on individual tasks. Similarly, Valindria {\it et al.} \cite{Valindria:2018} employed a MTL framework to improve the performance for multi-organ segmentation from CT and MR images, exploring various encoder-decoder network architectures. Specific to the cardiac MR applications, Xue {\it et al.} \cite{Xue:2018} proposed a network capable of learning multi-task relationship in a Bayesian framework to estimate various local/global LV indices for full quantification of the LV. Similarly, Dangi {\it et al.} \cite{Dangi:2019} performed joint segmentation and quantification of the LV myocardium using the learned task uncertainties to weigh the losses, improving upon the state-of-the-art results. Most of these MTL methods in medical image analysis aim to perform various clinically relevant tasks simultaneously. However, the focus of this work is on improving the segmentation performance of various FCN architectures using MTL as a network regularizer.

We propose to use the rich information available in the distance map of the segmentation mask as an auxiliary task for the image segmentation network. Since each pixel in the distance map represents its distance from the closest object boundary, this representation is redundant and robust compared to the per-pixel image label used for semantic segmentation. Furthermore, the distance map represents the shape and boundary information of the object to be segmented. Hence, training the segmentation network on the additional task of predicting the distance map is equivalent to enforcing shape and boundary constraints for the segmentation task; hence the name distance map regularized convolutional neural network.

Related work to ours include \cite{Bai:2017}, which take an image and its semantic segmentation as input and predict the distance transform of the object instances, such that, thresholding the distance map yields the instance segmentation. Similarly, \cite{Hayder:2017} represent the boundary of the object instances using a truncated distance map, which is used to refine the instance segmentation result. However, unlike these methods, our goal is not to perform instance segmentation, but to refine the semantic segmentation result using the distance map as an auxiliary task. The most closely related work to ours is presented in \cite{Bischke:2017} for segmentation of building footprints from satellite images using a MTL framework. 
In their study, the truncated distance map is predicted at the end of the decoder network and is further used to refine the boundary of the predicted segmentation, resulting in increased model complexity. 
% and high sensitivity of the output segmentation to the distance map threshold.
% Their study is limited to a binary segmentation task, and their network architecture results in increased model complexity. 
% Whereas, in this work, we perform both binary and multi-class segmentations and propose a generic framework to use MTL as a network regularizer, without increasing the model complexity.
Unlike that work, we impose a global shape constraint at the bottleneck layer of FCN architectures, using MTL as a network regularizer without increasing the model complexity. 
% The proposed architecture reduces the sensitivity of the output segmentation to the distance map threshold,
The proposed model is customized towards cardiac MRI image segmentation, as we accommodate for slices containing no foreground pixels (in apical and basal regions). Furthermore, we demonstrate better generalization performance of the proposed network with improved cross-dataset segmentation results.

\begin{figure*}[!th]
\centering
\includegraphics[width=\textwidth]{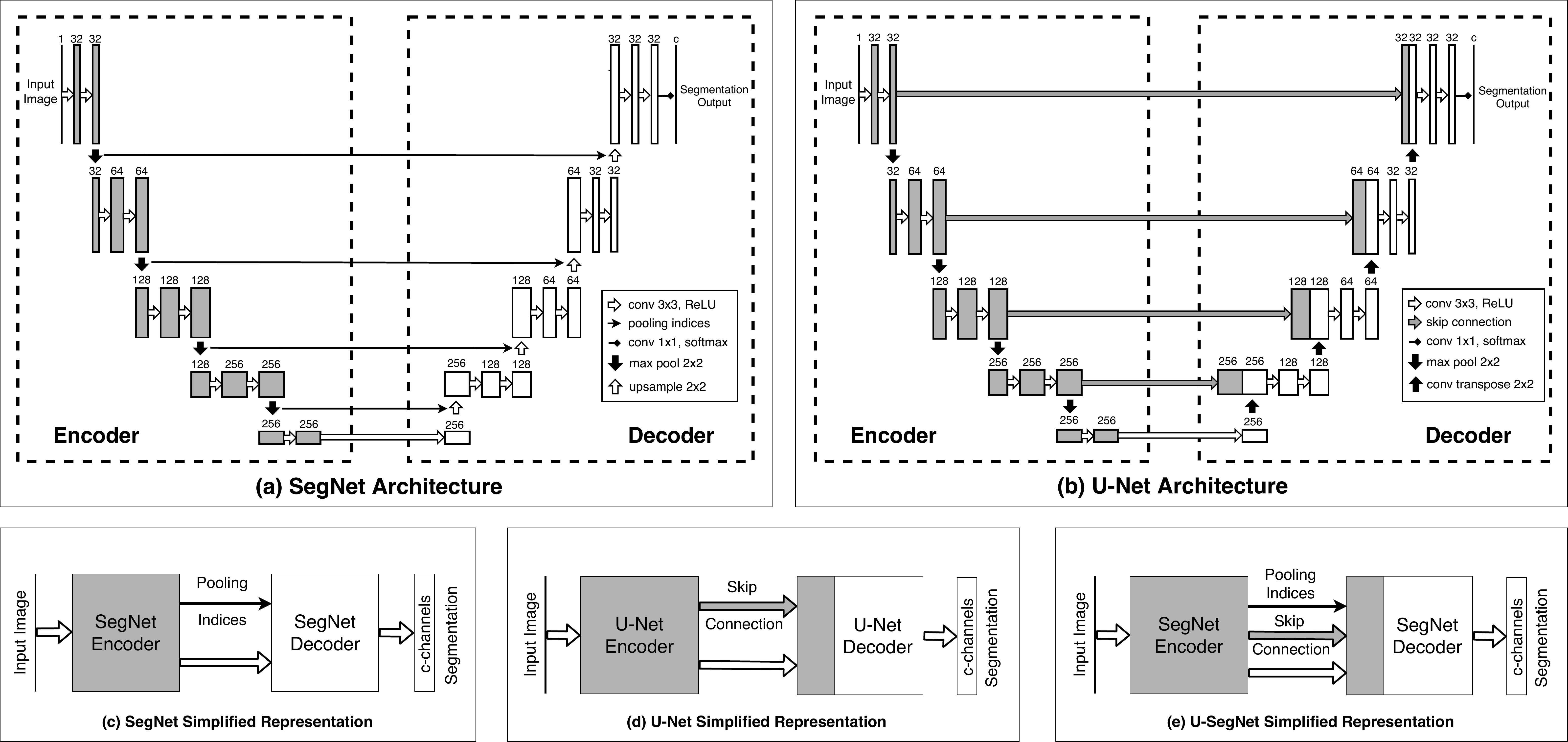}
\caption{Baseline FCN architectures and their simplified block representation.}
\label{fig:NetworkArchitectures}
\vspace{-5mm}
\end{figure*}

{\it Contributions: }
In this work, we propose to impose shape and boundary constraints in a CNN framework to accurately segment the heart chambers from cardiac cine MR images. We impose soft-constraints by including a distance map prediction as an auxiliary task in a MTL framework. We extensively evaluate our proposed model on two publicly available cardiac cine MRI datasets. We demonstrate that the addition of a distance map regularization block improves the segmentation performance of three FCN architectures, without increasing the model complexity and inference time. We employ a task uncertainty-based weighing scheme to automatically learn the weights for the segmentation and distance map regression tasks during training, and show that this method improves segmentation performance over the fixed equal-weighting scheme. Additionally, we show that the proposed regularization technique improves the segmentation performance in the challenging apical and basal slices, as well as across several different pathological heart conditions. This improvement is also reflected on the computed clinical indices important for cardiac health diagnosis. Finally, we demonstrate better generalization ability using the proposed regularization technique with significantly improved cross-dataset segmentation performance, without tuning the network to a new data distribution.

% \subsection*{Contributions}
% The main contributions of this work are:
% \begin{itemize}
% % \vspace{2mm}
% \item We use distance map prediction as an auxiliary task in a MTL framework to impose soft-constraints on the shape and boundary of the objects to be segmented.
% \vspace{1mm}
% % \item We demonstrate the application of the proposed regularization method for binary as well as multi-class segmentation on two different cardiac MRI datasets.
% \vspace{1mm}
% \item We demonstrate that the addition of a distance map regularization block improves the segmentation performance of three FCN architectures without increasing the model complexity and inference time. 
% \vspace{1mm}
% \item We apply the task uncertainty-based weighing scheme to automatically learn weights for the segmentation and distance map regression tasks during training.
% % \vspace{0mm}
% \item We demonstrate superior generalization ability of the proposed regularization technique with significantly improved cross-dataset segmentation performance, without tuning the network to a new data distribution.
% \end{itemize}

\section{Methods and Materials}
% In this section, we describe two popular FCN architectures for image segmentation. We then present our distance map regularization block and show how it can be incorporated into a existing FCN architecture. We then summarize the task-uncertainty based weighting of segmentation and regression losses. Finally, we present quantitative metrics used to evaluate the segmentation performance and the estimated clinical indices.

\subsection{CNN for Semantic Image Segmentation}
Let $\vec{x} = \{x_i \in {\rm I\!R}, i \in \mathcal{S}\}$ be the input intensity image and $\vec{y} = \{y_i \in \mathcal{L}, i \in \mathcal{S}\}$ be the corresponding image segmentation, with $\mathcal{C}=\{0,1,2,...,C-1\}$ representing a set of $C$ class labels, and $\mathcal{S}$ representing the image domain. The task of a CNN based segmentation model, with weights $\vec{W}$, is to learn a discriminative function $\vec{f}^{\vec{W}}(\cdot)$ that models the underlying conditional probability distribution 
$p(\vec{y}|\vec{x})$. The output of a CNN model is passed through a softmax function to produce a probability distribution over the class labels, such that, the function $\vec{f}^{\vec{W}}(\cdot)$ can be learned by maximizing the likelihood:
% \begin{equation}
% \begin{split}
%     p(\vec{y}=c|\vec{f}^{\vec{W}}(\vec{x})) & = \text{Softmax}(\vec{f}_c^{\vec{W}}(\vec{x})) \\
%     & = \frac{\text{exp}\left(\vec{f}_{c}^{\vec{W}}(\vec{x})\right)}{\sum_{c'\in \mathcal{L}}\text{exp}\left(\vec{f}_{c'}^{\vec{W}}(\vec{x})\right)}
% \end{split}
% \label{eq:seglikelihood}
% \end{equation}
\begin{equation}
\small
    p(\vec{y}=c|\vec{f}^{\vec{W}}(\vec{x})) = \text{Softmax}(\vec{f}_c^{\vec{W}}(\vec{x}))  = \frac{\text{exp}\left(\vec{f}_{c}^{\vec{W}}(\vec{x})\right)}{\sum_{c'\in \mathcal{L}}\text{exp}\left(\vec{f}_{c'}^{\vec{W}}(\vec{x})\right)}
\label{eq:seglikelihood}
\end{equation}
where $\vec{f}_{c}^{\vec{W}}(\vec{x})$ represents the $c$'th element of the vector $\vec{f}^{\vec{W}}(\vec{x})$. In practice, the negative log-likelihood $-\text{log}(p(\vec{y}|\vec{f}^{\vec{W}}(\vec{x})))$ is minimized to learn the optimal CNN model weights, $\vec{W}$. This is equivalent to minimizing the cross-entropy loss of the ground-truth segmentation, $\vec{y}$, with respect to the softmax of the network output, $\vec{f}^{\vec{W}}(\vec{x})$.

A typical FCN architecture ({\bf Fig. \ref{fig:NetworkArchitectures}}) for image segmentation consists of an encoder and a decoder network. The encoder network includes multiple pooling (max/average pooling) layers applied after several convolution and non-linear activation layers (e.g. Rectified linear unit (ReLU) \cite{Krizhevsky:2012}). It encodes hierarchical features important for the image segmentation task. To obtain per-pixel image segmentation, the global features obtained at the bottleneck layer need to be up-sampled to the original image resolution using the decoder network. The up-sampling filters can either be fixed (e.g. nearest-neighbor or bilinear upsampling), or can be learned during the training (deconvolutional layer). The final output of a decoder network is passed to a softmax classifier to obtain a per-pixel classification.

In a SegNet \cite{Badrinarayanan:2015} ({\bf Fig. \ref{fig:NetworkArchitectures}a}) architecture, the decoder produces sparse feature maps by up-sampling its inputs using the pooling indices transferred from its encoder. These sparse feature maps are then convolved with a trainable filter bank to obtain dense feature maps, and are finally passed through a softmax classifier to produce per-pixel image segmentation. Since the decoder in the SegNet architecture uses only the global features obtained at the bottleneck layer of the encoder, the high frequency details in the segmentation are lost during the up-sampling process. 

The U-Net architecture \cite{Ronneberger:2015} ({\bf Fig. \ref{fig:NetworkArchitectures}b}) introduced skip connections, by concatenating output of encoder layers at different resolutions to the input of the decoder layers at corresponding resolutions, hence preserving the high frequency details important for accurate image segmentation. Furthermore, the skip connections are known to ease the network optimization \cite{He:2016CVPR} by introducing multiple paths for backpropagation of the gradients, hence, mitigating the vanishing/exploding gradient problem. Similarly, skip connections also allow the network to learn lower level details in the outer layers and focus on learning the residual global features in the deeper encoder layers. Hence, the U-Net architecture is able to produce excellent segmentation results using limited training data with augmentation, and has been extensively used in medical image segmentation.

We observed that learned deconvolution filters in the original U-Net architecture can be replaced by a SegNet-like decoder
% , i.e. up-sampling the feature maps using the pooling indices transferred from the encoder,
to form a hybrid architecture with reduced network parameters.
% The up-sampled sparse features are then densified using convolution filters and concatenated with the feature maps obtained from the skip connection. We hypothesize this architecture, using the encoder pooling indices during up-sampling, ensures most information flows through the bottleneck layer, such that, the proposed regularization on the bottleneck layer has higher effect. This simplified up-sampling path facilitates the network training, while also reducing the network parameters. 
We refer to this modified architecture as U-SegNet ({\bf Fig \ref{fig:NetworkArchitectures}e}) throughout this paper, and use it as one of the baseline FCN architectures.

%\IEEEpubidadjcol

\begin{figure}[!h]
\centering
\includegraphics[width=0.48\textwidth]{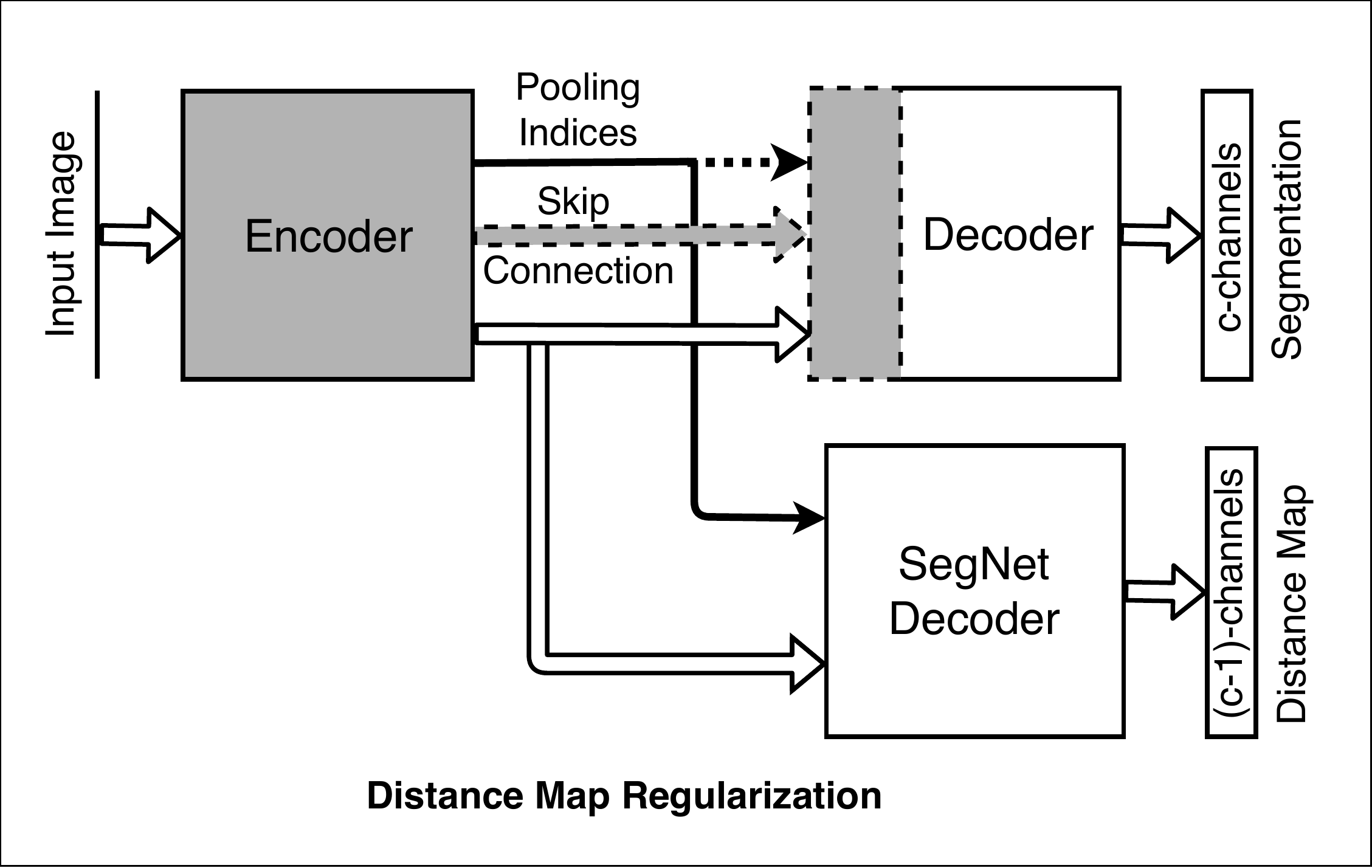}
\caption{Distance map regularizer added to the bottleneck layer. The number of distance map channels is one (1) fewer than the number of classes. Segmentation networks optionally use the pooling indices (yes/no) and skip-connections (yes/no), shown by dashed lines, during decoding: (a) DMR-SegNet: pooling indices (yes), skip connections (no); (b) DMR-USegNet: pooling indices (yes), skip connections (yes); and (c) DMR-UNet: pooling indices (no), skip connections (yes).}
% Segmentation networks optionally use the pooling indices and skip-connections (shown by dashed lines) during decoding, (a) DMR-SegNet: yes, no; (b) DMR-UNet: no, yes; and (c) DMR-USegNet: yes, yes; respectively. }
\label{fig:DistMapBlock}
\vspace{-5mm}
\end{figure}

\subsection{Distance Map Regularization Network}
The distance map of a binary segmentation mask can be obtained by computing the Euclidean distance of each pixel from the nearest boundary pixel \cite{Borgefors:1986}. This representation provides rich, redundant, and robust information about the boundary, shape, and location of the object to be segmented. For a binary segmentation mask, where $\Omega = \{x_i : y_i=1, i\in\mathcal{S}\}$ is the set of foreground pixels, $\partial \Omega$ represent the boundary pixels, and $d(\cdot,\cdot)$ is the Euclidean distance between any two pixels, the truncated signed distance map, $D(\vec{x})$, is computed as:
\begin{equation}
\begin{aligned}
D(x_i) = 
\begin{cases}
	d(x_i,\partial \Omega) & \text{if}~x_i \in \Omega, \Omega \notin \O\\
	-\text{min}(d(x_{i},\partial \Omega), T) & \text{if}~x_i \notin \Omega, \Omega \notin \O\\
	-T & \text{if}~\Omega \in \O
\end{cases}\\
\end{aligned}
\label{eq:DistanceMap}
\end{equation}
where,
\begin{equation*}
    d(x_i,\partial \Omega) = \min_{q_i \in \partial \Omega} d(x_i,q_i)
\end{equation*}
is the minimum distance of pixel $x_i \in \vec{x}$ from the boundary pixels $q_i \in \partial \Omega$. We truncate the signed distance map at a predefined distance threshold, $-T$, hence assigning this maximum negative distance to the slices not containing any foreground pixels (i.e. $\Omega \in \O$), indicating all pixels in the slice are far from the foreground (typically in the apical/basal regions of cardiac cine MR images).

The distance map regularization network is a SegNet-like decoder network, up-sampling the feature maps obtained at the bottleneck layer of the encoder to the size of the input image, with the number of output channels equal to the number of foreground classes (i.e. $C-1$). For example, for a four-class segmentation problem ($C=4$): background, RV blood-pool, LV myocardium, and LV blood-pool, the regularization network has three output channels, predicting the truncated signed distance maps ({Eq.~\ref{eq:DistanceMap}}) computed from the binary masks of the foreground classes: RV bood-pool, LV myocardium, and LV blood-pool. 

{\bf Fig. \ref{fig:DistMapBlock}} shows the regularization network added to the bottleneck layer of existing FCN architectures. Network training loss is the weighted sum of the cross-entropy loss for segmentation and the mean absolute difference (MAD) loss between the predicted and the reference distance maps.
% Although the predicted distance maps are similar to the ground-truth maps, we observed they are not accurate and hence cannot be used to refine the obtained segmentation. 
% Since the MAD error between the predicted and reference distance map is large (shown in supplementary materials {\bf Fig.~S2}), we remove the regularization block after training, such that, the original FCN architecture remains unchanged.
Since our goal is to perform semantic segmentation we do not need the distance map prediction at inference time. Therefore, we remove the regularization block after training, such that, the original FCN architecture remains unchanged. Additionally, we found that the quality (mean absolute difference) of the predicted distance maps is insufficient for improving the predicted segmentations from the standard path (see {\bf Fig.~S2} in supplement).

\subsection{MTL using Uncertainty-based Loss Weighting}
\label{subsec:taskweighting}

We model the likelihood for a segmentation task as the squashed and scaled version of the model output through a softmax function:
\begin{equation}
p(\vec{y}|\vec{f}^{\vec{W}}(\vec{x}),\sigma) = \text{Softmax}\left(\frac{1}{\sigma^2}\vec{f}^{\vec{W}}(\vec{x})\right)
\end{equation}
where, $\sigma$ is a positive scalar, equivalent to the {\it temperature}, for the defined Gibbs/Boltzmann distribution. The magnitude of $\sigma$ determines how {\it uniform} the discrete distribution is, and hence relates to the uncertainty of the prediction measured in entropy. The log-likelihood for the segmentation task can be written as:
\begin{equation}
\begin{split}
& \text{log}~p(\vec{y}=c|\vec{f}^{\vec{W}}(\vec{x}),\sigma)\\
&~~~~~~~~~~~~~ = \frac{1}{\sigma^2}f_{c}^{\vec{W}}(\vec{x})-\text{log}\sum_{c'}\text{exp}\left(\frac{1}{\sigma^2}f_{c'}^{\vec{W}}(\vec{x})\right)\\
&~~~~~~~~~~~~~ = \frac{1}{\sigma^2}(f_{c}^{\vec{W}}(\vec{x})-\text{log}\sum_{c'}\text{exp}\left(f_{c'}^{\vec{W}}(\vec{x})\right)- \hfill\\
&~~~~~~~~~~~~~~~~~~~~ \text{log}\frac{\sum_{c'}\text{exp}\left(\frac{1}{\sigma_2^2}f_{c'}^{\vec{W}}(\vec{x})\right)}{\left(\sum_{c'}\text{exp}\left(f_{c'}^{\vec{W}}(\vec{x})\right)\right)^{\frac{1}{\sigma_2^2}}}\\
&~~~~~~~~~~~~~ \approx \frac{1}{\sigma^2}\text{log}~\text{Softmax}\left(\vec{y},\vec{f}^{\vec{W}}(\vec{x})\right) - \text{log} \sigma
\end{split}
\end{equation}
where $f_{c}^{\vec{W}}(\vec{x})$ is the $c$'th element of the vector $\vec{f}^{\vec{W}}(\vec{x})$. In the last step, a simplifying assumption $\frac{1}{\sigma}\sum_{c'}\text{exp}\left(\frac{1}{\sigma^2}f_{c'}^{\vec{W}}(\vec{x})\right) \approx {\left(\sum_{c'}\text{exp}\left(f_{c'}^{\vec{W}}(\vec{x})\right)\right)^{\frac{1}{\sigma^2}}}$, which becomes an equality when $\sigma \rightarrow 1$, has been made, resulting in a simple optimization objective with improved empirical results \cite{Kendall:2017MTL}.

Similarly, for the regression task, we define our likelihood as a Lapacian distribution with its mean and scale parameter given by the neural network output:
\begin{equation}
p(\vec{y}|\vec{f}^{\vec{W}}(\vec{x}),\sigma) = \frac{1}{2\sigma}\text{exp}\left(-\frac{|\vec{y}-\vec{f}^{\vec{W}}(\vec{x})|}{\sigma}\right)
% \mathcal{N}(\vec{f}^{\vec{W}}(\vec{x}),\sigma^2)
\end{equation}
The log-likelihood for regression task can be written as:
\begin{equation}
\text{log}~p(\vec{y}|\vec{f}^{\vec{W}}(\vec{x}),\sigma) \approx -\frac{1}{\sigma}|\vec{y}-\vec{f}^{\vec{W}}(\vec{x})|-\text{log}\sigma
\end{equation}
where $\sigma$ is the neural networks observation noise parameter --- capturing the noise in the output. A constant term has been removed for simplicity, as it does not affect the optimization.

For a network with two outputs --- continuous output $\vec{y}_1$ modeled with a Laplacian likelihood, and a discrete output $\vec{y}_2$ modeled with a softmax likelihood --- the joint loss is:
\begin{equation}
\begin{split}
& \mathcal{L}(\vec{W_1,W_2},\sigma_1,\sigma_2) \\
& = -\text{log}~p(\vec{y}_1,\vec{y}_2=c|\vec{f}^{\vec{W_1}}(\vec{x}),\vec{f}^{\vec{W_2}}(\vec{x}),\sigma_1,\sigma_2) \\
& = -\text{log}~(p(\vec{y}_1|\vec{f}^{\vec{W_1}}(\vec{x}),\sigma_1) \cdot p(\vec{y}_2=c|\vec{f}^{\vec{W_2}}(\vec{x}),\sigma_2))\\ 
% & = \frac{1}{\sigma_1} |\vec{y_{1}}-\vec{f}^{\vec{W}}(\vec{x})|+ \text{log} \sigma_1 + \frac{1}{\sigma_2^2} \mathcal{L}_2(\vec{W})+ \\
% & ~~~~~~~~~~~~\text{log}\frac{\sum_{c'}\text{exp}\left(\frac{1}{\sigma_2^2}f_{c'}^{\vec{W}}(\vec{x})\right)}{\left(\sum_{c'}\text{exp}\left(f_{c'}^{\vec{W}}(\vec{x})\right)\right)^{\frac{1}{\sigma_2^2}}}\\
& \approx \frac{1}{\sigma_1} \mathcal{L}_1(\vec{W_1}) + \frac{1}{\sigma_2^2} \mathcal{L}_2(\vec{W_2})+ \text{log} \sigma_1 + \text{log} \sigma_2\\
\end{split}
\label{eq:jointloss}
\end{equation}
where $\mathcal{L}_1(\vec{W_1})=|\vec{y_{1}}-\vec{f}^{\vec{W_1}}(\vec{x})|$ is the MAD loss of $\vec{y}_1$ and $\mathcal{L}_2(\vec{W_2})=-\text{log}~\text{Softmax}(\vec{y}_2,\vec{f}^{\vec{W_2}}(\vec{x}))$ is the cross-entropy loss of $\vec{y}_2$. To arrive at {Eq. \ref{eq:jointloss}}, the two tasks are assumed independent.
% and simplifying assumption $\frac{1}{\sigma}\sum_{c'}\text{exp}\left(\frac{1}{\sigma^2}f_{c'}^{\vec{W}}(\vec{x})\right) \approx {\left(\sum_{c'}\text{exp}\left(f_{c'}^{\vec{W}}(\vec{x})\right)\right)^{\frac{1}{\sigma^2}}}$, which becomes an equality when $\sigma \rightarrow 1$, has been made for the softmax likelihood, resulting in a simple optimization objective with improved empirical results \cite{Kendall:2017MTL}. 
During the training, the joint likelihood loss $\mathcal{L}(\vec{W_1},\vec{W_2},\sigma_1,\sigma_2)$ is optimized with respect to $\vec{W_1}$, $\vec{W_2}$ as well as $\sigma_1$, $\sigma_2$.

From {Eq. \ref{eq:jointloss}}, we can observe that the losses for individual tasks are weighted by the inverse of their corresponding uncertainties ($\sigma_1$, $\sigma_2$) learned during the training. Hence, the task with higher uncertainty will be weighted less and vice versa. Furthermore, the uncertainties cannot grow too large due to the penalty imposed by the last two terms in {(Eq. \ref{eq:jointloss})}. In practice, the network is trained to predict the log variance, $s:=\text{log}\sigma$, for numerical stability and avoiding any division by zero, such that, the positive scale parameter, $\sigma$, can be computed via exponential mapping $\text{exp}(s)$.

\subsection{Clinical Datasets}

\subsubsection{Left Ventricle Segmentation Challenge (LVSC)}
This study employed 97 de-identified cardiac MRI image datasets from patients suffering from myocardial infraction and impaired LV contraction available as a part of the STACOM 2011 Cardiac Atlas Segmentation Challenge project \cite{Fonseca:2011,Suinesiaputra:2014} database\footnote{\url{http://www.cardiacatlas.org/challenges/lv-segmentation-challenge/}}. Cine-MRI images in short-axis and long-axis views are available for each case. The images were acquired using the Steady-State Free Precession (SSFP) MR imaging protocol with the following settings: typical thickness $\leq 10mm$, gap $\leq 2mm$, TR $30-50ms$, TE $1.6ms$, flip angle $60 ^0$, FOV $360mm$, spatial resolution $0.7031$ to $2.0833$ $mm^2/pixel$ and $256\times256mm$ image matrix using multiple scanners from various manufacturers. Corresponding reference myocardium segmentation generated from expert analyzed 3D surface finite element model are available for all 97 cases throughout the cardiac cycle.

\subsubsection{Automated Cardiac Diagnosis Challenge (ACDC)}
This dataset\footnote{\url{https://www.creatis.insa-lyon.fr/Challenge/acdc/databases.html}} is composed of short-axis cardiac cine-MR images acquired for 100 patients divided into 5 evenly distributed subgroups: normal, myocardial infarction, dilated cardiomyopathy, hypertropic cardiomyopathy, and abnormal right ventricle, available as a part of the STACOM 2017 ACDC challenge  \cite{Bernard:2018}. The acquisitions were obtained over a 6 year period using two MRI scanners of different magnetic strengths (1.5T and 3.0T). The images were acquired using the SSFP sequence with the following settings: thickness $5mm$ (sometimes $8mm$), interslice gap $5mm$, spatial resolution $1.37$ to $1.68$ $mm^2/pixel$, $28$ to $40$ frames per cardiac cycle. Corresponding manual segmentations for RV blood-pool, LV myocardium, and LV blood-pool, performed by a clinical expert for the end-systole (ES) and end-diastole (ED) phases are provided.

\subsection{Data Preprocessing and Augmentation}
SimpleITK \cite{Yaniv:2018} was used to resample short-axis images to a common resolution of 1.5625 $mm^2/pixel$ and crop/zero-pad to a common size of $192\times192$ and $256\times256$ for LVSC and ACDC dataset, respectively. Image intensities were clipped at 99$^{th}$ percentile and normalized to zero mean and unit standard deviation. Each dataset was divided into $80\%$ train, $10\%$ validation, and $10\%$ test set with five non-overlaping folds for cross-validation. Train-validation-test fold was performed randomly over the whole LVSC dataset, whereas it was performed per subgroup (stratified sampling) for the ACDC dataset to maintain even distribution of subgroups over the training, validation, and testing sets. The training images were subjected to random similarity transform with: isotropic scaling of $0.8$ to $1.2$, rotation of $0^{o}$ to $360^{o}$, and translation of $-1/8^{th}$ to $+1/8^{th}$ of the image size along both x- and y-axes. The training set for LVSC and ACDC dataset included the original images along with augmentation of two and four randomly transformed versions of each image, respectively. We heavily augment the ACDC dataset, as the labels are available only for the ES and ED phases, whereas, lightly augment the LVSC dataset, as the labels are available throughout the cardiac cycle.

\subsection{Network Training and Testing Details}
Networks implemented in PyTorch\footnote{https://github.com/pytorch/pytorch} were initialized with the {\it Kaiming uniform} initializer \cite{He:2015} and trained for 30 and 100 epochs for LVSC and ACDC dataset, respectively, with batch size of 15 images. {\it RMS prop} optimizer \cite{Hinton:2012neural} with a learning rate of 0.0001 and 0.0005 for single- and multi-task networks, respectively, decayed by 0.99 every epoch was used. We saved the model with best average Dice coefficient on the validation set, and evaluated on the test set. 
% The best performing network, in terms of the average Dice coefficient between the obtained and reference segmentation, in the validation set, was used for evaluation on the test set.

Networks were trained on NVIDIA Titan Xp GPU. The distance map threshold was selected empirically and set to a large value of $250$ $pixels$, i.e. full distance map. The cross-entropy and the MAD loss were initialized with equal weights of $1.0$, such that, the optimal weighting was learned automatically. The auxillary task of distance map regression was removed after the network training. The obtained 2D slice segmentations were rearranged into a 3D volume, and the largest connected component for each heart chamber was retained to yield the final segmentation. Model complexity and average timing requirements for training and testing the models is shown in {\bf Table \ref{tab:TrainTest}}. 

\scriptsize
\begin{table}[ht]
{\caption{Model complexity, training and testing time. The model size for DMR networks are equivalent to corresponding baseline FCN architectures during test time. The inference time for DMR networks without removing the regularization block are shown in brackets.}
\label{tab:TrainTest}}
\begin{center}
\vspace{-2mm}
\begin{tabular}{|p{1.8cm}||>{\centering\arraybackslash}m{0.6cm}|>{\centering\arraybackslash}m{0.6cm}|>{\centering\arraybackslash}m{0.7cm}|>{\centering\arraybackslash}m{0.7cm}||>{\centering\arraybackslash}m{0.5cm}|>{\centering\arraybackslash}m{0.5cm}|}
\hline
\multirow{2}{1.8cm} & \multicolumn{4}{>{\centering\arraybackslash}p{2.6cm}||}{Time } & \multicolumn{2}{>{\centering\arraybackslash}p{1.0cm}|}{\multirow{3}{1.0cm}{\#Parameters ($\times 10^6$)}}\\ 
\cline{2-5}
\multirow{2}{1.8cm} & \multicolumn{2}{>{\centering\arraybackslash}p{1.2cm}|}{Train (min/epoch)} & \multicolumn{2}{>{\centering\arraybackslash}p{1.4cm}||}{Test (ms/volume)} & \multicolumn{2}{>{\centering\arraybackslash}p{1.0cm}|}{}\\ 
\cline{2-7}
& ACDC & LVSC & ACDC & LVSC & Train & Test \\
\hline \hline 
SegNet & 2.49 & 14.91 & 70 & 67 & 2.96 & 2.96 \\
\hline 
USegNet & 2.41 & 14.49 & 70 & 67 & 3.75 & 3.75 \\
\hline 
UNet & 2.65 & 15.50 & 72 & 68 & 4.10 & 4.10 \\
\hline 
DMR-SegNet & 4.44 & 20.57 & 70(157) & 63(94) & 3.56 & 2.96 \\
\hline 
DMR-USegNet & 4.84 & 19.03 & 73(158) & 65(96) & 4.35 & 3.75 \\
\hline
DMR-UNet & 4.85 & 21.16 & 75(160) & 67(97) & 4.70 & 4.10 \\
\hline 
\end{tabular}
\end{center}
\vspace{-5mm}
\end{table}
\normalsize

\subsection{Evaluation Metrics}
We use overlap and surface distance measures to evaluate the segmentation. Additionally, we evaluate the clinical indices associated with the segmentation.

\subsubsection{Dice and Jaccard Coefficients}
Given two binary segmentation masks, A and B, the Dice and Jaccard coefficient are defined as:

\begin{equation}
    \text{Dice} = \frac{2|A \cap B|}{|A|+|B|},~~~\text{Jaccard} = \frac{|A \cap B|}{|A \cup B|}
\end{equation}

where, $| \cdot |$ gives the cardinality (i.e. the number of non-zero elements) of each set. Maximum and minimum values (1.0 and 0.0, repectively) for Dice and Jaccard coefficient occur when there is 100\% and 0\% overlap between the two binary segmentation masks, respectively.

\subsubsection{Mean Surface Distance and Hausdorff Distance}
Let, $S_A$ and $S_B$, be surfaces (with $N_A$ and $N_B$ points, respectively) corresponding to two binary segmentation masks, A and B, respectively. The mean surface distance (MSD) is defined as:
\begin{equation}
    \text{MSD} = \frac{1}{2}\left(\frac{1}{N_A}\sum_{p \in S_A}d(p,S_B)+\frac{1}{N_B}\sum_{q \in S_B}d(q,S_A)\right)
\end{equation}
Similarly, Hausdorff Distance (HD) is defined as:
\begin{equation}
    \text{HD} = \text{max}\left(\max_{p \in S_A}d(p,S_B),\max_{q \in S_B}d(q,S_A)\right)
\end{equation}
where, 
\begin{equation*}
    d(p,S) = \min_{q \in S} d(p,q)
\end{equation*}
is the minimum Euclidean distance of point $p$ from the points $q \in S$. Hence, MSD computes the mean distance between the two surfaces, whereas, HD computes the largest distance between the two surfaces, and is sensitive to outliers.

\subsubsection{Ejection Fraction and Myocardial Mass}
Ejection Fraction (EF) is an important cardiac parameter quantifying the cardiac output. EF is defined as:
\begin{equation}
    \text{EF} = \frac{\text{EDV}-\text{ESV}}{\text{EDV}}\times 100 \%
\end{equation}
where, EDV is the end-diastolic volume, and ESV is the end-systolic volume. Similarly, the myocardial mass can be computed from the myocardial volume as:
\begin{equation}
    \text{Myo-Mass} = \text{Myo-Volume}~(cm^3)\times 1.06~(gram/cm^3)
\end{equation}
The correlation coefficients for the EF and myocardial mass computed from the ground-truth versus those computed from the automatic segmentation is reported. Correlation coefficient of $+1$ ($-1$) represents perfect positive (negative) linear relationship, whereas that of $0$ represents no linear relationship between two variables.

\subsubsection{Limits of Agreement} To compare the clinical indices computed from the ground-truth versus those obtained from the automatic segmentation, we take the difference between each pair of the two observations. The mean of these differences is termed as {\it bias}, and the 95\% confidence interval, mean $\pm $1.96$ \times $standard deviation (assuming a Gaussian distribution), is termed as {\it limits of agreement} (LoA).

\section{Results}

\scriptsize
\begin{table*}[ht]
{\caption{Evaluation of the average segmentation results on ACDC dataset for RV blood-pool, LV myocardium, and LV blood-pool (mean value reported), obtained from all networks against the provided reference segmentation. The statistical significance of the results for DM regularized model compared against the baseline model are represented by $^{*}$ and $^{**}$ for p-values less than $0.05$ and $0.005$, respectively. Also shown are the clinical indices evaluated for each heart chamber. The best performing model for each metric has been {\bf highlighted}. SN: SegNet, USN: USegNet, UNet: UNet.}
\label{tab:SegmentationACDC}}
\begin{subtable}[t]{\textwidth}
\caption{Evaluation of Average (across all heart chambers) Segmentation Results}
\vspace{-3mm}
\label{tab:RVSegmentation}
\begin{center}
\begin{tabular}{|p{1.4cm}||>{\centering\arraybackslash}m{0.9cm}>{\centering\arraybackslash}m{0.9cm}|>{\centering\arraybackslash}m{0.9cm}>{\centering\arraybackslash}m{0.9cm}|>{\centering\arraybackslash}m{0.9cm}>{\centering\arraybackslash}m{0.9cm}||>{\centering\arraybackslash}m{0.9cm}>{\centering\arraybackslash}m{0.9cm}|>{\centering\arraybackslash}m{0.9cm}>{\centering\arraybackslash}m{0.9cm}|>{\centering\arraybackslash}m{0.9cm}>{\centering\arraybackslash}m{0.9cm}|}

\hline
\multirow{2}{1.4cm} & \multicolumn{6}{c||}{End Diastole (ED)} & \multicolumn{6}{|c|}{End Systole (ES)}\\ 
\cline{2-13}
& SN & DMR SN & USN & DMR USN & UNet & DMR UNet & SN & DMR SN & USN & DMR USN & UNet & DMR UNet \\
\hline \hline
% Dice (\%) & 91.1 (2.9) & 91.7$^{*}$ (2.7) & 91.5 (3.0) & 92.0$^{**}$ (2.7) & 91.6 (2.5) & {\bf92.2$^{**}$} (3.0) & 87.3 (4.9) & 88.0 (5.4) & 87.7 (5.1) & {\bf88.7$^{**}$} (5.0) & 87.2 (7.1) & 88.8$^{*}$ (4.8)\\
% \hline
% Jaccard (\%) & 84.0 (4.4) & 85.1$^{*}$ (4.2) & 84.7 (4.7) & 85.5$^{**}$ (4.3) & 85.0 (4.0) & {\bf85.9$^{**}$} (4.7) & 78.1 (7.0) & 79.3 (7.6) & 78.7 (7.5) & {\bf80.3$^{**}$} (7.4) & 78.3 (9.3) & 80.4$^{*}$ (7.2) \\
% \hline
% MSD (mm) & 0.55 (0.34) & 0.53 (0.49) & 0.58 (0.56) & {\bf0.52$^{*}$} (0.46) & 0.54 (0.30) & 0.53$^{*}$ (0.57) & 0.92 (0.59) & 0.85 (0.66) & 0.92 (0.61) & {\bf0.84} (0.67) & 1.08 (1.24) & 0.83 (0.65) \\
% \hline
% HD (mm) & 10.26 (4.08) & 9.87 (4.77) & 10.26 (4.53) & 9.67 (4.19) & 10.03 (3.93) & {\bf9.52} (5.03) & 11.33 (3.72) & 10.31 (4.17) & 11.66 (4.09) & {\bf10.91} (4.66) & 12.61 (5.49) & 10.96$^{*}$ (4.47) \\
Dice (\%) & 91.1 & 91.7$^{**}$ & 91.5 & 92.0$^{**}$ & 91.6 & {\bf92.2$^{**}$} & 87.3 & 88.0$^{*}$ & 87.7 & 88.7$^{**}$ & 87.2 & {\bf88.8$^{*}$} \\
\hline
Jaccard (\%) & 84.0 & 85.1$^{**}$ & 84.7 & 85.5$^{**}$ & 85.0 & {\bf85.9$^{**}$} & 78.1 & 79.3$^{*}$ & 78.7 & 80.3$^{**}$ & 78.3 & {\bf80.4$^{*}$} \\
\hline
MSD (mm) & 0.55 & 0.53$^{*}$ & 0.58 & {\bf0.52$^{*}$} & 0.54 & 0.53$^{*}$ & 0.92 & 0.85 & 0.92 & 0.84 & 1.08 & {\bf0.83} \\
\hline
HD (mm) & 10.26 & 9.87 & 10.26 & 9.67 & 10.03 & {\bf9.52} & 11.33 & {\bf10.31$^{*}$} & 11.66 & 10.91 & 12.61 & 10.96$^{*}$ \\

\hline
\end{tabular}
\end{center}
\end{subtable}

\vspace{2mm}

\begin{subtable}[t]{\textwidth}
\caption{Evaluation of the Clinical Indices}
\vspace{-3mm}
\label{tab:RVSegmentation}
\begin{center}
\begin{tabular}{|p{1.4cm}||>{\centering\arraybackslash}m{0.9cm}>{\centering\arraybackslash}m{0.9cm}|>{\centering\arraybackslash}m{0.9cm}>{\centering\arraybackslash}m{0.9cm}|>{\centering\arraybackslash}m{0.9cm}>{\centering\arraybackslash}m{0.9cm}||>{\centering\arraybackslash}m{0.9cm}>{\centering\arraybackslash}m{0.9cm}|>{\centering\arraybackslash}m{0.9cm}>{\centering\arraybackslash}m{0.9cm}|>{\centering\arraybackslash}m{0.9cm}>{\centering\arraybackslash}m{0.9cm}|}

\hline
\multirow{2}{1.4cm} & \multicolumn{6}{c||}{Correlation Coefficient} & \multicolumn{6}{|c|}{Bias+LOA}\\ 
\cline{2-13}
& SN & DMR SN & USN & DMR USN & UNet & DMR UNet & SN & DMR SN & USN & DMR USN & UNet & DMR UNet \\
\hline \hline
LV EF & 0.939 & 0.947 & 0.944 & {\bf0.970} & 0.962 & 0.963 & 1.00 (13.15) & 0.31 (12.44) & 0.58 (12.57) & -0.42 {\bf(9.24)} & 0.31 (10.41) & 0.40 (10.40) \\
\hline
RV EF & 0.874 & 0.871 & 0.866 & {\bf0.895} & 0.856 & 0.870 & 1.04 (17.40) & 1.77 (17.34) & 0.85 (17.40) & 0.38 {\bf(15.42)} & 0.09 (18.94) & 0.29 (18.30) \\
\hline
Myo Mass & 0.948 & 0.970 & 0.958 & 0.973 & 0.933 & {\bf0.978} & 3.10 (32.94) & -0.43 (25.17) & 0.35 (29.65) & 0.21 (23.89) & 2.85 (37.39) & 0.80 {\bf(21.75)} \\
\hline
\end{tabular}
\end{center}
\end{subtable}

\end{table*}
\normalsize

\scriptsize
\begin{table*}[ht]
{\caption{Comparison of the segmentation results obtained from the DMR-UNet model against the top three ACDC challenge participants, evaluated on the held-out 50 patient challenge testset. The Dice metric, Hausdorff Distance (HD), and correlation of clinical indices for all three heart chambers is shown.}
\vspace{-3mm}
\label{tab:SegmentationACDCChallenge}}
\begin{center}
\begin{tabular}{|p{1.8cm}||>{\centering\arraybackslash}m{0.5cm}>{\centering\arraybackslash}m{0.5cm}|>{\centering\arraybackslash}m{0.5cm}>{\centering\arraybackslash}m{0.5cm}|>{\centering\arraybackslash}m{0.5cm}>{\centering\arraybackslash}m{0.5cm}>{\centering\arraybackslash}m{0.6cm}||>{\centering\arraybackslash}m{0.5cm}>{\centering\arraybackslash}m{0.5cm}|>{\centering\arraybackslash}m{0.5cm}>{\centering\arraybackslash}m{0.5cm}|>{\centering\arraybackslash}m{0.5cm}>{\centering\arraybackslash}m{0.5cm}>{\centering\arraybackslash}m{0.6cm}||>{\centering\arraybackslash}m{0.6cm}|>{\centering\arraybackslash}m{0.6cm}|}
\hline
\multirow{2}{1.8cm} & \multicolumn{7}{c||}{End Diastole (ED)} & \multicolumn{7}{|c||}{End Systole (ES)} & \multicolumn{2}{c|}{EF}\\ 
\cline{2-17}
\multirow{2}{1.8cm} & \multicolumn{2}{c|}{LV} & \multicolumn{2}{c|}{RV} & \multicolumn{3}{c||}{Myo} & \multicolumn{2}{c|}{LV} & \multicolumn{2}{c|}{RV} & \multicolumn{3}{c||}{Myo} & LV & RV\\ 
\cline{2-17}
& Dice & HD & Dice & HD & Dice & HD & Corr & Dice & HD & Dice & HD & Dice & HD & Corr & Corr & Corr\\
\hline \hline
Baumgartner\cite{Baumgartner:2018} & 0.96 & 6.53 & 0.93 & 12.67 & 0.89 & 8.70 & 0.982 & 0.91 & 9.17 & 0.88 & 14.69 & 0.90 & 10.64 & 0.983 & 0.988 & 0.851 \\
\hline
Khened\cite{Khened:2018} & 0.96 & 8.13 & 0.94 & 13.99 & 0.89 & 9.84 & {\bf0.990} & 0.92 & 8.97 & 0.88 & 13.93 & 0.90 & 12.58 & 0.979 & 0.989 & 0.858 \\
\hline
Isensee\cite{Isensee:2018} & {\bf0.97} & 7.38 & {\bf0.95} & 10.12 & {\bf0.90} & 8.72 & 0.989 & {\bf0.93} & {\bf6.91} & {\bf0.90} & {\bf12.14} & {\bf0.92} & 8.67 & 0.985 & {\bf0.991} & {\bf0.901} \\
\hline
{\bf DMR-UNet} & 0.96 & {\bf6.05} & 0.94 & {\bf9.52} & 0.89 & {\bf7.92} & 0.989 & 0.92 & 8.16 & 0.88 & 13.05 & 0.91 & {\bf8.39} & {\bf0.987} & 0.989 & 0.851 \\
\hline
\end{tabular}
\end{center}
\vspace{-3mm}
\end{table*}
\normalsize

\scriptsize
\begin{table*}[ht]
{\caption{Evaluation of the segmentation results on LVSC dataset for LV myocardium (mean values reported), obtained from all networks against the provided reference segmentation. The statistical significance of the results for DM regularized model compared against the baseline model are represented by $^{*}$ and $^{**}$ for p-values less than $0.05$ and $0.005$, respectively. The best performing model for each metric has been highlighted. SN: SegNet, USN: USegNet, UNet: UNet.}
\vspace{-3mm}
\label{tab:SegmentationLVSC}}
\begin{center}
\begin{tabular}{|p{1.4cm}||>{\centering\arraybackslash}m{0.9cm}>{\centering\arraybackslash}m{0.9cm}|>{\centering\arraybackslash}m{0.9cm}>{\centering\arraybackslash}m{0.9cm}|>{\centering\arraybackslash}m{0.9cm}>{\centering\arraybackslash}m{0.9cm}||>{\centering\arraybackslash}m{0.9cm}>{\centering\arraybackslash}m{0.9cm}|>{\centering\arraybackslash}m{0.9cm}>{\centering\arraybackslash}m{0.9cm}|>{\centering\arraybackslash}m{0.9cm}>{\centering\arraybackslash}m{0.9cm}|}

\hline
\multirow{2}{1.4cm} & \multicolumn{6}{c||}{End Diastole (ED)} & \multicolumn{6}{|c|}{End Systole (ES)}\\ 
\cline{2-13}
& SN & DMR SN & USN & DMR USN & UNet & DMR UNet & SN & DMR SN & USN & DMR USN & UNet & DMR UNet \\
\hline \hline
% Dice (\%) & 82.2 (4.1) & 83.0$^{*}$ (3.7) & 82.5 (4.0) & 83.2$^{*}$ (3.8) &  83.1 (3.9) & {\bf83.6} (4.0) & 83.5 (4.5) & 84.2 (4.0) & 83.8 (4.0) & 84.3 (4.2) & 84.3 (4.1) & {\bf84.6} (4.0) \\
% \hline
% Jaccard (\%) & 70.0 (5.7) & 71.1$^{*}$ (5.2) & 70.4 (5.5) & 71.5$^{*}$ (5.3) & 71.3 (5.5) & {\bf72.0} (5.6) & 71.9 (6.5) & 72.9 (5.8) & 72.4 (5.8) & 73.0 (6.1) & 73.0 (6.0) & {\bf73.5} (5.8) \\
% \hline
% MSD (mm) & 0.78 (0.28) & 0.74 (0.26) & 0.79 (0.28) & 0.72 (0.27) & 0.74 (0.28) & {\bf0.70} (0.29) & 0.81 (0.33) & 0.77 (0.28) & 0.77 (0.29) & 0.78 (0.32) & 0.74 (0.28) & {\bf0.75} (0.28)\\
% \hline
% HD (mm) & 13.20 (3.55) & 13.14 (3.87) & 13.67 (3.79) & 13.12 (4.03) & 12.98 (4.04) & {\bf12.80} (4.08) & 12.96 (3.75) & 12.96 (3.57) & 12.71$^{*}$ (3.71) & 13.71 (4.44) & 13.08 (3.68) & {\bf12.51} (3.06) \\
Dice (\%) & 82.2 & 83.0$^{*}$ & 82.5 & 83.2$^{**}$ &  83.1 & {\bf83.6} & 83.5 & 84.2 & 83.8 & 84.3$^{*}$ & 84.3 & {\bf84.6} \\
\hline
Jaccard (\%) & 70.0 & 71.1$^{*}$ & 70.4 & 71.5$^{**}$ & 71.3 & {\bf72.0} & 71.9 & 72.9 & 72.4 & 73.0$^{*}$ & 73.0 & {\bf73.5} \\
\hline
MSD (mm) & 0.78 & 0.74 & 0.79 & 0.72$^{*}$ & 0.74 & {\bf0.70} & 0.81 & 0.77 & 0.77 & 0.78 & {\bf0.74} & 0.75 \\
\hline
HD (mm) & 13.20 & 13.14 & 13.67 & 13.12 & 12.98 & {\bf12.80} & 12.96 & 12.96 & 12.71$^{*}$ & 13.71 & 13.08 & {\bf12.51} \\
\hline
Mass (Corr) & 0.908 & 0.937 & 0.923 & {\bf0.938} & 0.917 & 0.936 & 0.921 & 0.935 & 0.929 & 0.926 & {\bf0.939} & 0.922\\
\hline
Mass(gram) (Bias+LOA) & 2.56 (35.25) & -0.92 (29.52) & 3.91 (32.48) & 3.34 {\bf(29.08)} & 2.88 (33.75) & 0.06 (29.92) & 5.48 (32.49) &  1.96 (29.58) & 5.04 (30.92) & 5.49 (31.49) & 5.18 {\bf(28.79)} & 2.56 (32.18) \\
\hline
\end{tabular}
\end{center}
\vspace{-4mm}
\end{table*}
\normalsize

\scriptsize
\begin{table*}[ht]
{\caption{Comparison of the LV myocardium segmentation results on the LVSC validation set against the consensus segmentation ({\bf CS$^*$}) as described in \cite{Suinesiaputra:2014}. The values for AU, AO, SCR, and INR are obtained from Table 2 in \cite{Suinesiaputra:2014}, CNR from Table 3 in \cite{Tan:2017}, FCN from Table 3 in \cite{Tran:2016}, and DFCN from Table 12 in \cite{Khened:2018}. Values are provided as mean (standard deviation), and in descending order by Jaccard index. SA/FA --- Semi/Fully-Automatic}
\vspace{-3mm}
\label{tab:ChallengeLVSC}}
\begin{center}
\begin{tabular}{|p{1.5cm}||>{\centering\arraybackslash}m{0.8cm}|>{\centering\arraybackslash}m{1.3cm}|>{\centering\arraybackslash}m{1.3cm}|>{\centering\arraybackslash}m{1.3cm}|>{\centering\arraybackslash}m{1.3cm}|>{\centering\arraybackslash}m{1.3cm}|}

\hline
Method & SA/FA & Jaccard & Sensitivity & Specificity & PPV & NPV \\
\hline \hline
AU \cite{Li:2010} & SA & 0.84 (0.17) & 0.89 (0.13) & 0.96 (0.06) & 0.91 (0.13) & 0.95 (0.06) \\
\hline
CNR \cite{Tan:2017} & SA & 0.77 (0.11) & 0.88 (0.09) & 0.95 (0.04) & 0.86 (0.11) & 0.96 (0.02) \\
\hline
FCN \cite{Tran:2016} & FA & 0.74 (0.13) & 0.83 (0.12) & 0.96 (0.03) & 0.86 (0.10) & 0.95 (0.03) \\
\hline
DFCN \cite{Khened:2018} & FA & 0.74 (0.15) & 0.84 (0.16) & 0.96 (0.03) & 0.87 (0.10) & 0.95 (0.03) \\
\hline
{\bf DMR-UNet} & FA & 0.74 (0.16) & 0.85 (0.16) & 0.95 (0.03) & 0.86 (0.10) & 0.95 (0.03) \\
\hline
AO \cite{Fahmy:2012} & SA & 0.74 (0.16) & 0.88 (0.15) & 0.91 (0.06) & 0.82 (0.12) & 0.94 (0.06) \\
\hline
SCR \cite{Jolly:2012} & FA & 0.69 (0.23) & 0.74 (0.23) & 0.96 (0.05) & 0.87 (0.16) & 0.89 (0.09) \\
\hline
INR \cite{Margeta:2012} & FA & 0.43 (0.10) & 0.89 (0.17) & 0.56 (0.15) & 0.50 (0.10) & 0.93 (0.09) \\
\hline
\end{tabular}
\end{center}
\vspace{-4mm}
\end{table*}
\normalsize

\subsection{Segmentation and Clinical Indices Evaluation}
The proposed Distance Map Regularized (DMR) SegNet, USegNet, and UNet models along with the baseline models were trained for the joint segmentation of RV blood-pool, LV myocardium, and LV blood-pool from the ACDC challenge dataset. The provided reference segmentation and the corresponding automatic segmentation obtained from the DMR-UNet model for a test patient is shown in {\bf Fig. \ref{fig:SegmentationResults}}. Automatic segmentation obtained from all networks, for ED and ES phases, are evaluated against the reference segmentation and summarized in {\bf Table \ref{tab:SegmentationACDC}a}; also shown is the evaluation of subsequently computed clinical indices in {\bf Table \ref{tab:SegmentationACDC}b}.

We observe consistent improvement in the average segmentation performance of the models after the DM-Regularization. Specifically, there is statistically significant improvement\footnote{Wilcoxon signed-rank test is performed for statistical significance testing} on several segmentation metrics for all evaluated models. Same results manifest onto the clinical indices with better correlation and LoA on both EF and myocardium mass. Furthermore, the DMR-UNet model outperforms other evaluated networks in many segmentation metrics.
% The proposed regularization in the bottleneck layer is more effective on the SegNet and USegNet models, as the decoder uses encoder indices to upsample the images, such that, most information flows through the bottleneck layer. However, since the deconvolution filters are learned in the UNet architecture, the information need not necessarily flow through the bottleneck layer, rendering the regularization less effective. 
% Furthermore, the DMR-UsegNet model consistently outperforms all other evaluated networks, with statistically significant improvement in the Dice and Jaccard coefficients for RV blood-pool and LV myocardium in both ED and ES phases. 

\begin{figure}[htb]
\centering
\begin{subfigure}[b]{0.48\textwidth}
  \vspace{-2mm}
  \includegraphics[width=1\linewidth]{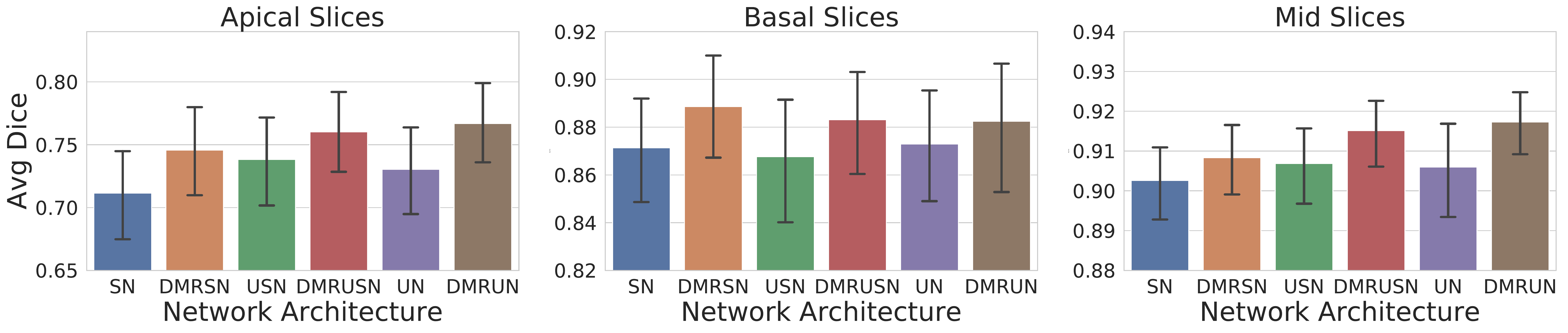}
  \caption{{\scriptsize Average Dice coefficient for LV blood-pool, LV myocardium, and RV blood-pool segmentation on ACDC dataset (100 volumes).}}
  \label{fig:DiceRegionalAll-A} 
\end{subfigure}
\par\bigskip
\begin{subfigure}[b]{0.48\textwidth}
  \vspace{-2mm}
  \includegraphics[width=1\linewidth]{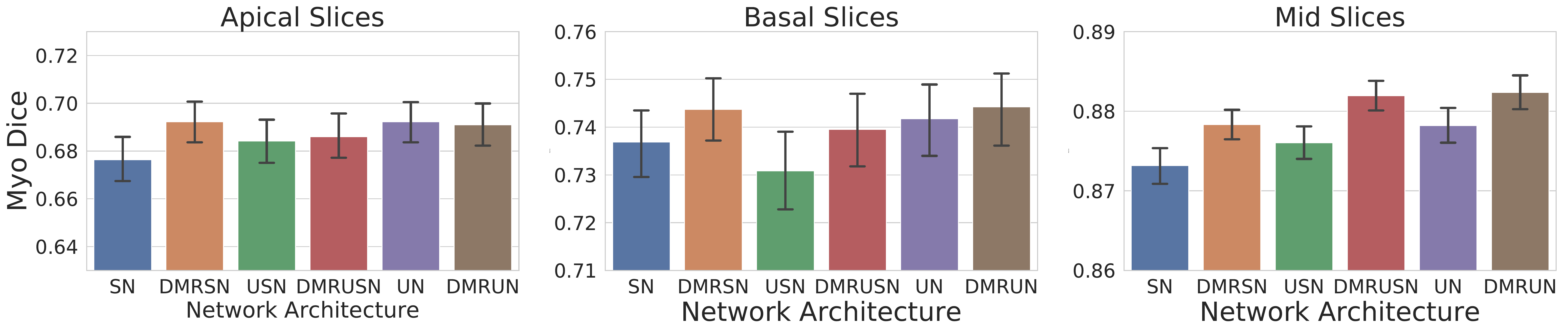}
  \caption{{\scriptsize Dice coefficient for myocardium segmentation on LVSC dataset (1050 volumes).}}
  \label{fig:DiceRegionalAll-B}
\end{subfigure}
\caption{Mean and $95\%$ bootstrap confidence interval for average Dice coefficient on apical, basal, and mid slices.}
% We can observe higher improvement in Dice coefficient for apical slices after the distance map regularization.}
\label{fig:DiceRegionalAll}
% \vspace{-4mm}
\end{figure}

\begin{figure}[htb]
\centering
\includegraphics[width=0.47\textwidth]{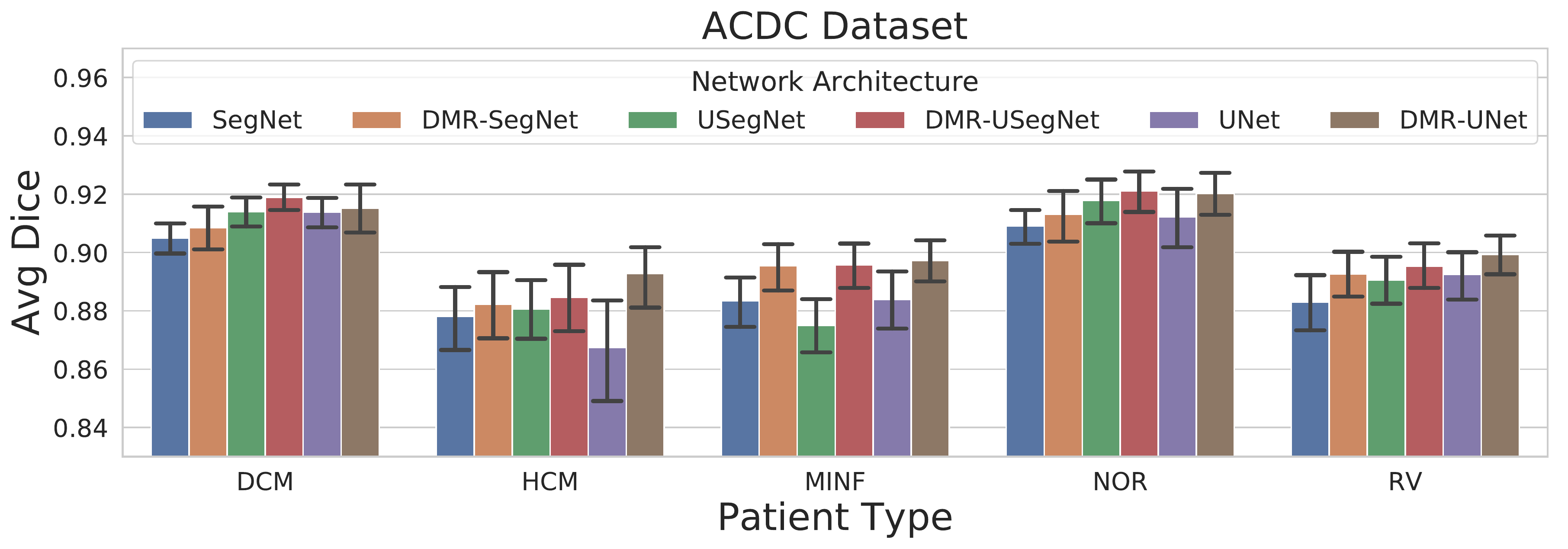}
\caption{Mean and $95\%$ bootstrap confidence interval of average Dice coefficient for segmentation results on ACDC dataset obtained from several architectures divided according to the five sub-groups: DCM --- dilated cardiomyopathy, HCM --- hypertrophic cardiomyopathy, MINF --- previous myocardial infarction, NOR --- normal subjects, and RV --- abnormal right ventricle.}
% We can observe consistent improvement in segmentation performance after the distance map regularization for most patient sub-groups.}
\label{fig:ComparePatients}
% \vspace{-4mm}
\end{figure}

\scriptsize
\begin{table*}[htb]
{\caption{Cross-dataset segmentation evaluation for LV myocardium segmentation (mean values reported). The statistical significance of the results for DM regularized model compared against the baseline model are represented by $^{*}$ and $^{**}$ for p-values less than $0.01$ and $0.001$, respectively. SN: SegNet, USN: USegNet, UNet: UNet.}
\label{tab:CrossDataset}}
\begin{subtable}[t]{\textwidth}
\caption{Trained on ACDC and tested on LVSC (194 volumes)}
\label{tab:ACDCtoLVSC}
\begin{center}
\vspace{-2mm}
\begin{tabular}{|p{1.4cm}||>{\centering\arraybackslash}m{0.9cm}>{\centering\arraybackslash}m{0.9cm}|>{\centering\arraybackslash}m{0.9cm}>{\centering\arraybackslash}m{0.9cm}|>{\centering\arraybackslash}m{0.9cm}>{\centering\arraybackslash}m{0.9cm}||>{\centering\arraybackslash}m{0.9cm}>{\centering\arraybackslash}m{0.9cm}|>{\centering\arraybackslash}m{0.9cm}>{\centering\arraybackslash}m{0.9cm}|>{\centering\arraybackslash}m{0.9cm}>{\centering\arraybackslash}m{0.9cm}|}

\hline
\multirow{2}{1.4cm} & \multicolumn{6}{c||}{End Diastole (ED)} & \multicolumn{6}{|c|}{End Systole (ES)}\\ 
\cline{2-13}
& SN & DMR SN & USN & DMR USN & UNet & DMR UNet & SN & DMR SN & USN & DMR USN & UNet & DMR UNet \\
\hline \hline
% Dice (\%) & 70.4 (13.0) & 73.3$^{**}$ (11.1) & 68.3 (15.6) & {\bf76.6$^{**}$} (9.9) & 72.3 (13.4) & 76.7$^{**}$ (10.2) & 68.0 (16.3) & 71.9$^{**}$ (16.2) & 65.5 (17.6) & 74.9$^{**}$ (14.8) & 69.7 (16.8) & {\bf76.4$^{**}$} (12.4) \\
% \hline
% Jaccard (\%) & 55.6 (12.8) & 58.9$^{**}$ (11.4) & 53.6 (14.6) & 62.9$^{**}$ (10.2) & 58.0 (13.1) & {\bf63.1$^{**}$} (10.8) & 53.3 (15.3) & 58.1$^{**}$ (15.4) & 50.8 (16.6) & 61.5$^{**}$ (14.4) & 55.5 (16.0) & {\bf63.1$^{**}$} (12.5) \\
% \hline
% MSD (mm) & 2.68 (3.24) & {\bf2.07$^{**}$} (3.08) & 3.33 (4.47) & 1.80$^{**}$ (3.38) & 2.46 (3.49) & 1.80$^{**}$ (3.10) & 3.56 (4.92) & 2.93$^{**}$ (4.64) & 4.19 (5.64) & 2.58$^{**}$ (4.58) & 3.49 (5.35) & {\bf2.35$^{**}$} (4.82) \\
% \hline
% HD (mm) & 25.01 (10.53) & 22.44$^{**}$ (10.54) & 26.93 (12.25) & {\bf20.33$^{**}$} (10.03) & 24.61 (10.38) & 20.16$^{**}$ (10.83) & 25.96 (12.18) & 22.62^{**}$ (12.43) & 27.37 (13.02) & {\bf21.67$^{**}$} (13.14) & 25.68 (12.01) & 20.98$^{**}$ (11.95) \\
Dice (\%) & 70.4 & 73.3$^{**}$ & 68.3 & 76.6$^{**}$ & 72.3 & {\bf76.7$^{**}$} & 68.0 & 71.9$^{**}$ & 65.5 & 74.9$^{**}$ & 69.7 & {\bf76.4$^{**}$} \\
\hline
Jaccard (\%) & 55.6 & 58.9$^{**}$ & 53.6 & 62.9$^{**}$ & 58.0 & {\bf63.1$^{**}$} & 53.3 & 58.1$^{**}$ & 50.8 & 61.5$^{**}$ & 55.5 & {\bf63.1$^{**}$} \\
\hline
MSD (mm) & 2.68 & 2.07$^{**}$ & 3.33 & 1.80$^{**}$ & 2.46 & {\bf1.80$^{**}$} & 3.56 & 2.93$^{**}$ & 4.19 & 2.58$^{**}$ & 3.49 & {\bf2.35$^{**}$} \\
\hline
HD (mm) & 25.01 & 22.44$^{**}$ & 26.93 & 20.33$^{**}$ & 24.61 & {\bf20.16$^{**}$} & 25.96 & 22.62$^{**}$ & 27.37 & 21.67$^{**}$ & 25.68 & {\bf20.98$^{**}$} \\
\hline
\end{tabular}
\end{center}
\end{subtable}

\vspace{3mm}
\begin{subtable}[t]{\textwidth}
\caption{Trained on LVSC and tested on ACDC (200 volumes)}
\label{tab:LVSCtoACDC}
\begin{center}
\vspace{-2mm}
\begin{tabular}{|p{1.4cm}||>{\centering\arraybackslash}m{0.9cm}>{\centering\arraybackslash}m{0.9cm}|>{\centering\arraybackslash}m{0.9cm}>{\centering\arraybackslash}m{0.9cm}|>{\centering\arraybackslash}m{0.9cm}>{\centering\arraybackslash}m{0.9cm}||>{\centering\arraybackslash}m{0.9cm}>{\centering\arraybackslash}m{0.9cm}|>{\centering\arraybackslash}m{0.9cm}>{\centering\arraybackslash}m{0.9cm}|>{\centering\arraybackslash}m{0.9cm}>{\centering\arraybackslash}m{0.9cm}|}

\hline
\multirow{2}{1.4cm} & \multicolumn{6}{c||}{End Diastole (ED)} & \multicolumn{6}{|c|}{End Systole (ES)}\\ 
\cline{2-13}
& SN & DMR SN & USN & DMR USN & UNet & DMR UNet & SN & DMR SN & USN & DMR USN & UNet & DMR UNet \\
\hline \hline
% Dice (\%) & 69.5 (21.5) & 78.4$^{**}$ (14.9) & 62.5 (23.9) & {\bf80.1$^{**}$} (12.6) & 62.1 (25.3) & 80.2$^{**}$ (13.9) & 57.7 (27.9) & 77.6$^{**}$ (14.3) & 51.9 (29.5) & {\bf79.3$^{**}$} (12.6) & 50.3 (30.1) & 79.1$^{**}$ (13.2) \\
% \hline
% Jaccard  & 56.5 (20.6) & 66.3$^{**}$ (15.6) & 49.3 (22.5) & {\bf68.2$^{**}$} (14.0) & 49.3 (23.7) & 68.5$^{**}$ (14.5) & 45.4 (25.4) & 65.3$^{**}$ (16.2) & 40.1 (26.1) & {\bf67.3$^{**}$} (15.1) & 38.8 (26.5) & 67.1$^{**}$ (15.6) \\
% \hline
% MSD (mm) & 4.92 (10.55) & 1.77$^{**}$ (2.86) & 6.75 (14.19) & {\bf1.30$^{**}$} (2.14) & 6.29 (9.28) & 1.59$^{**}$ (3.79) & 9.59 (14.74) & 2.53$^{**}$ (3.74) & 13.27 (21.85) & {\bf2.35$^{**}$} (3.61) & 10.97 (12.62) & 2.52$^{**}$ (3.79) \\
% \hline
% HD (mm) & 26.04 (19.56) & 17.06$^{**}$ (15.12) & 29.08 (22.00) & {\bf13.93$^{**}$} (10.49) & 29.50 (18.77) & 14.16$^{**}$ (11.23) & 35.13 (24.35) & 19.25$^{**}$ (14.49) & 39.60 (29.79) & {\bf18.77$^{**}$} (15.54) & 37.44 (22.36) & 19.58$^{**}$ (15.26) \\
Dice (\%) & 69.5 & 78.4$^{**}$ & 62.5 & 80.1$^{**}$ & 62.1 & {\bf80.2$^{**}$} & 57.7 & 77.6$^{**}$ & 51.9 & {\bf79.3$^{**}$} & 50.3 & 79.1$^{**}$ \\
\hline
Jaccard  & 56.5 & 66.3$^{**}$ & 49.3 & 68.2$^{**}$ & 49.3 & {\bf68.5$^{**}$} & 45.4 & 65.3$^{**}$ & 40.1 & {\bf67.3$^{**}$} & 38.8 & 67.1$^{**}$ \\
\hline
MSD (mm) & 4.92 & 1.77$^{**}$ & 6.75 & {\bf1.30$^{**}$} & 6.29 & 1.59$^{**}$ & 9.59 & 2.53$^{**}$ & 13.27 & {\bf2.35$^{**}$} & 10.97 & 2.52$^{**}$ \\
\hline
HD (mm) & 26.04 & 17.06$^{**}$ & 29.08 & {\bf13.93$^{**}$} & 29.50 & 14.16$^{**}$ & 35.13 & 19.25$^{**}$ & 39.60 & {\bf18.77$^{**}$} & 37.44 & 19.58$^{**}$ \\
\hline
\end{tabular}
\end{center}
\end{subtable}
\vspace{-3mm}
\end{table*}
\normalsize

To further analyze the improvement in segmentation performance, we performed a regional analysis by sub-dividing the slices into apical (25\% slices in the apical region and beyond), basal (25\% slices in the basal region and beyond) and mid-region (remaining 50\% mid slices), based on the reference segmentation. From {\bf Fig. \ref{fig:DiceRegionalAll-A}}, we can observe consistent improvement in segmentation performance at the problematic apical and basal slices \cite{Bernard:2018}; however, due to the small size of these regions, the improvement does not have a large effect on the overall performance, though it is of significance when constructing patient specific models of the heart for simulation purposes \cite{Peters:2018}. We postulate that the additional constraint imposed by a very high negative distance assigned to empty apical/basal slices prevents the network from over-segmenting these regions, hence, improving the regional dice overlap and effectively reducing the overall Hausdorff distance.

To study the effect of the distance map regularization across the five patient sub-groups, we plot the average Dice coefficient for each sub-group computed for all six models in {\bf Fig. \ref{fig:ComparePatients}}. As expected, we observe the segmentation performance is better for the normal patients in comparison to the pathological cases. Furthermore, we observe consistent improvement in segmentation performance after the distance map regularization for all patient sub-groups.

% To compare the performance of our DM regularized U-Net model against the competing models from the ACDC challenge, 
We segmented the heart structures from 50 patients ACDC held-out testset and submitted to the challenge organizers. Majority voting prediction of ensemble of DMR-UNet models trained for five-fold cross-validation followed by a 3D connected component analysis yielded the final segmentation. {\bf Table \ref{tab:SegmentationACDCChallenge}} shows the comparison of our segmentation results against the top three methods submitted to the challenge. Baumgartner {\it et al.} \cite{Baumgartner:2018} tested several architectures and found that 2D U-Net with a cross-entropy loss performed the best. Khened {\it et al.} \cite{Khened:2018} used a 2D U-Net with dense blocks and an inception first layer to obtain the segmentation. Isensee {\it et al.} ensembled 2D and 3D U-Net architectures trained with a Dice loss to obtain the best result in the challenge. Our 2D DMR-UNet model is able to perform as good or better than the other two 2D methods, however, the combination of 2D and 3D context has marginal improvement in the Dice overlap metric. Based on this observation, we believe the ensemble of 2D and 3D DMR-UNet model should be able to perform as good or better than \cite{Isensee:2018}, which is not the main objective of this work. Nonetheless, we can observe the constraint imposed by the DM regularization is successful in reducing the errors in apical/basal regions, manifested in the improved Hausdorff distance.

{\bf Table \ref{tab:SegmentationLVSC}} shows the segmentation performance evaluated on the LVSC dataset, demonstrating superior performance of the DM regularized models over their baseline. Specifically, there is statistically significant improvement on the Dice and Jaccard metric for the ED phase. Furthermore, the correlation and LoA for the myocardial mass improves after network regularization. The improvement in performance is consistent across different heart regions as shown in {\bf Fig. \ref{fig:DiceRegionalAll-B}}.

We segmented the myocardium from the LVSC held-out validation set of 100 patients. Majority voting prediction from ensemble of DMR-UNet models trained for five-fold cross-validation followed by a 3D connected-component analysis yielded the final segmentation. {\bf Table \ref{tab:ChallengeLVSC}} shows our segmentation results (computed per slice) compared against several other semi-/fully-automatic algorithms. Reported segmentation results are computed against the consensus segmentation (CS$^*$) built from multiple challenge submissions \cite{Suinesiaputra:2014}. Segmentation results for the four challenge participants --- AU \cite{Li:2010}, AO \cite{Fahmy:2012}, SCR \cite{Jolly:2012}, and INR \cite{Margeta:2012}, and the details on segmentation evaluation metrics can be found in the challenge summary report \cite{Suinesiaputra:2014}. The AU method \cite{Li:2010} used the interactive guide-point modeling technique to fit a finite element cardiac model to the CMR data and required expert approval of all slices and all frames. This segmentation was provided as the reference segmentation to the challenge participants. The CNN regression (CNR) method \cite{Tan:2017} regressed the endo- and epi-cardium contours in polar coordinates, while manually eliminating the problematic slices beyond the apex and base of the heart, hence, obtaining a good segmentation result. The mean (std dev) of Jaccard coefficients computed for our DMR-UNet model in apical, mid, and basal slices are 0.66 (0.18), 0.77 (0.12), and 0.74 (0.17), respectively. Our DMR-UNet model has similar performance to competing fully-automatic segmentation algorithms based on the fully convolutional network (FCN) \cite{Tran:2016} and the densely connected FCN (DFCN) \cite{Khened:2018} architectures. The DFCN method involves a computationally expensive region of interest (ROI) identification based on a Fourier transform applied across the cardiac cycle, followed by the circular Hough transform; whereas our method requires minimal pre-processing. 
% Interestingly, the simple encoder-decoder SegNet architecture performs equivalent to the UNet architecture with skip connections, likely due to the large training dataset; also reflected in marginal improvement of segmentation performance on the ES phase after DM regularization. 

Lastly, the segmentation performance on the LVSC dataset ({\bf Table \ref{tab:ChallengeLVSC}}) is significantly lower than ACDC dataset ({\bf Table \ref{tab:SegmentationACDCChallenge}}) due to large variability and noise exhibited by the LVSC data as compared to the ACDC dataset.

\subsection{Cross Dataset Evaluation (Transfer Learning)}
To analyze the generalization ability of our proposed distance map regularized networks, we performed a cross-dataset segmentation evaluation. The networks trained on ACDC dataset for five-fold cross-validation were tested on the LVSC dataset, and vice versa; such that, the majority voting scheme produced the final per-pixel segmentation. 
% Quantitative evaluation of the automatic myocardium segmentation against the provided reference segmentation is summarized in {\bf Table \ref{tab:CrossDataset}}. 
We observe a significant boost in Dice coefficient of 5\% to 12\% for distance map regularized networks over their baseline models when trained on ACDC and tested on LVSC dataset (194 ED and ES volumes), as shown in {\bf Table \ref{tab:CrossDataset}a}. Similarly, the distance map regularized models significantly outperform the baseline models by 23\% to 42\% improvement in Dice coefficient, when trained on LVSC and tested on ACDC dataset (200 ED and ES volumes), as shown in {\bf Table \ref{tab:CrossDataset}b}. The improvement in generalization performance for the regularization networks trained on LVSC dataset is higher, likely due to the availability of large number of heterogeneous training examples. Similar improvement can be observed in the MSD and HD metric. We want to emphasize that our networks are trained separately on each dataset and are completely unaware of the new data distribution, unlike a typical domain adaptation \cite{Wang:2018} setting. Nonetheless, the distance map regularized networks are able to generalize better to a new dataset compared to the baseline models.

We further analyzed the feature maps across different layers of the baseline and distance map regularized networks (supplementary material {\bf Fig. S3}). We can observe the baseline models preserve the intensity information and propagate it throughout the network, hence, they are more sensitive to the dataset-specific intensity distribution. On the other hand, the multi-task regularized networks focus more on the edges and other discriminative features, producing sparse feature maps, while ignoring dataset-specific intensity distribution. Moreover, from the feature maps at the decoding layers, we observe a clear delineation of several cardiac structures in the regularized network, while those for the baseline models are less discriminative, and contain information about all structures present in the image. Hence, we verify that multi-task learning-based distance map regularization helps the network learn generalizable features important for the segmentation task, demonstrated by their excellent transfer learning capabilities (see Supplementary Materials for details on feature visualization ({\bf Fig. S3}) and network learning curves showing the robustness of distance map regularized models against overfitting ({\bf Fig. S4})).

\section{Discussion}

We performed an extensive study on the effects of hyper-parameters on the performance of the proposed regularization framework. Here we summarize the effects of the learned vs. fixed task weighting, and various choices of the distance map threshold. Furthermore, we analyzed the distribution of network weights before and after regularization.

\begin{figure}[htb]
\centering
\includegraphics[width=0.48\textwidth]{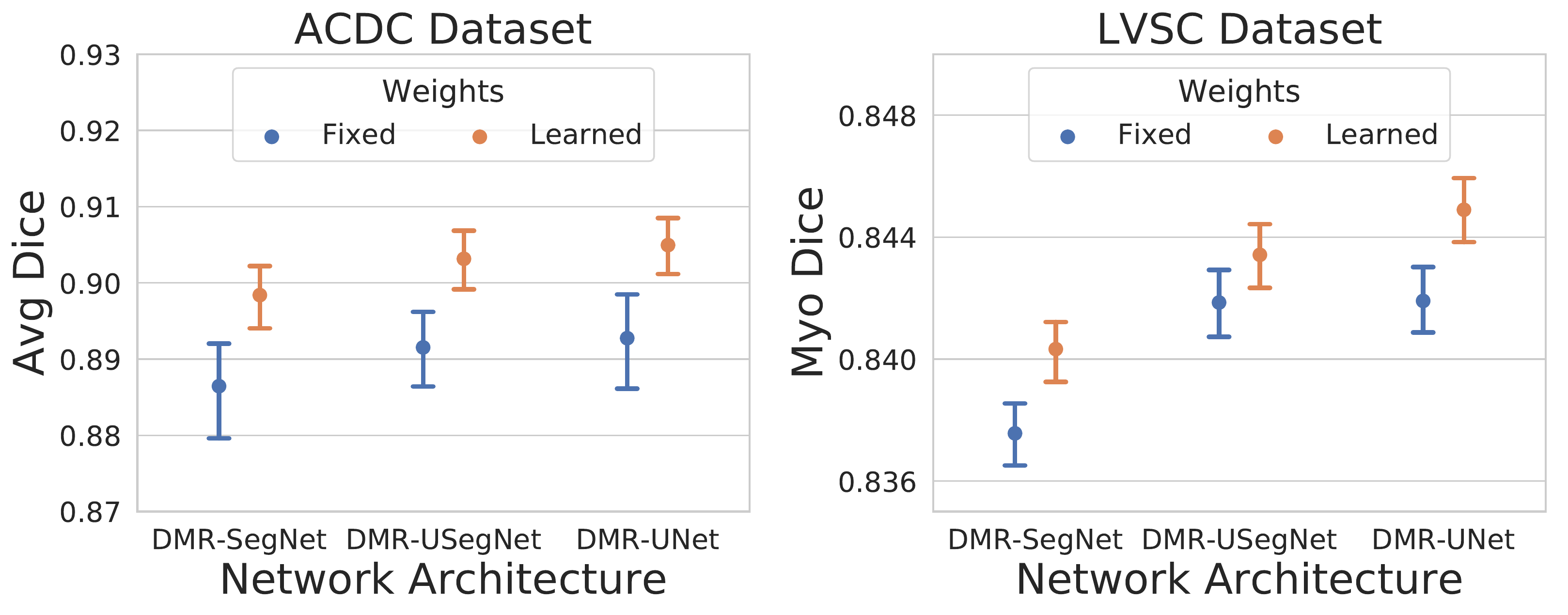}
\caption{Mean and $95\%$ bootstrap confidence interval of average Dice coefficient for Learned vs Fixed equal weighting. Learned task weighting statistically significantly improves the segmentation performance.}
\label{fig:WeightingComparison}
% \vspace{-4mm}
\end{figure}

% \noindent 
{\it Task Weighting: }
% \subsection{Task Weighting}
% During network training, we observed that the automatic weighting scheme learns to weigh the cross-entropy and MAD loss, such that, they are brought to the same scale. Hence, to accelerate network training, 
% We initialize the weights for the cross-entropy and MAD loss equally i.e. 1.0 and 1.0, respectively. The learned weights for (cross-entropy, MAD) are around (0.01, 17) and (0.02, 13) for ACDC and LVSC dataset, respectively, for the best performing models on the validation set. 
At first, we initialized the weights for the cross-entropy and MAD loss equally to 1.0. However, the learned weights for the cross-entropy and MAD loss were around 0.01 and 17, and 0.02 and 13 for ACDC and LVSC dataset, respectively, for the best performing models on the validation set. 

To determine the effect of learned task weighting scheme presented in section {\bf \ref{subsec:taskweighting}}, we analyzed the average Dice coefficient of the test set segmentation results for both ACDC (100 volumes) and LVSC (1050 volumes) datasets with fixed versus learned weighting. From {\bf Fig. \ref{fig:WeightingComparison}}, we can observe a significant improvement in average Dice coefficient (based on the 95\% bootstrap confidence intervals) with learned weights compared to fixed (equal) weighting. Since the scales of the two losses are different, the equal weighting scheme emphasizes the distance map regression task more than it should, hence deteriorating the segmentation performance. On the other hand, the learned task weighting scheme is able to automatically weigh the two losses, bringing them to a similar scale, such that the two tasks are given equal importance, ultimately improving the segmentation performance.

\begin{figure}[htb]
\centering
\includegraphics[width=0.48\textwidth]{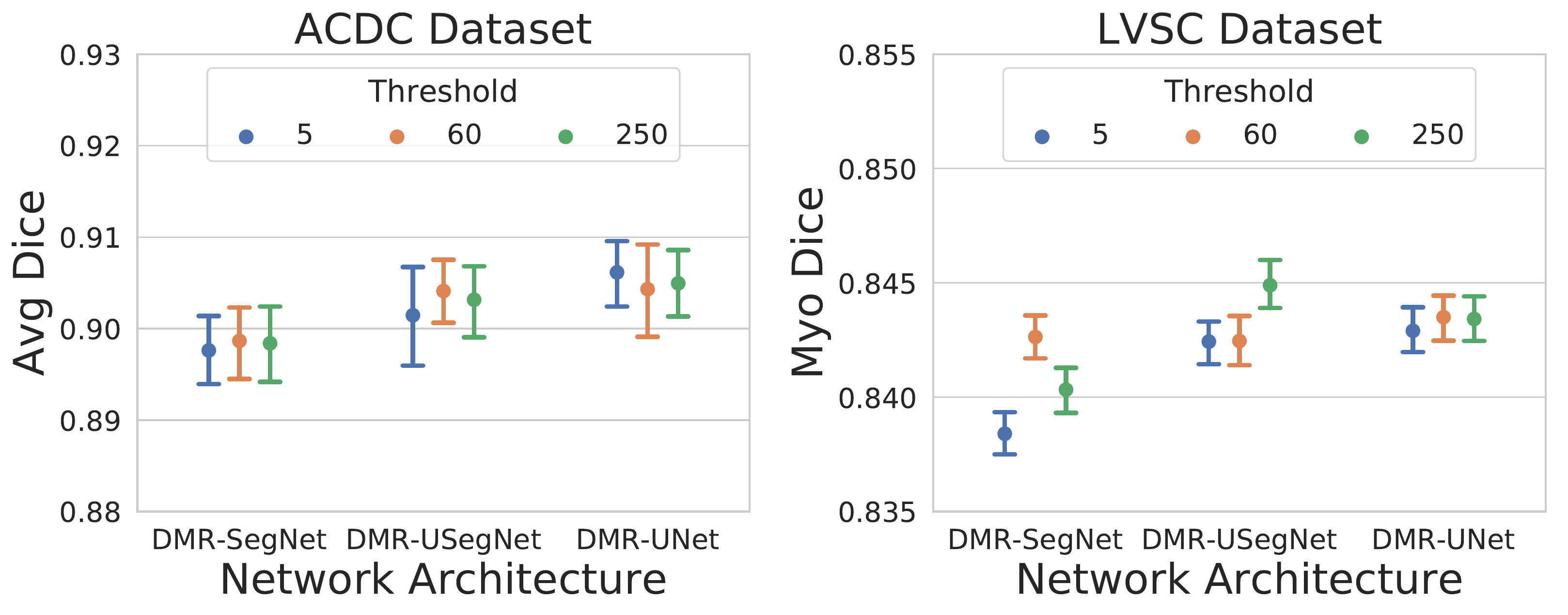}
\caption{Mean and $95\%$ bootstrap confidence interval of average Dice coefficient for a range of distance map thresholds.}
\label{fig:ThresholdComparison}
% \vspace{-4mm}
\end{figure}

% \noindent 
{\it Effect of Distance Map Threshold: }
We selected three extreme values for the distance map threshold: $5$, $60$, and $250$ $pixels$. The network weights for cross-entropy and MAD loss were equally initialized to $(1,1)$ and trained with automatically learned task weighting for a fixed number of epochs. The average Dice coefficient on the test-set obtained from the best performing models on the validation-set across five-fold cross-validation is summarized in {\bf Fig. \ref{fig:ThresholdComparison}}. We observe similar performance for different threshold values, demonstrating the low sensitivity of the proposed method to the distance map threshold.
% higher threshold values, with marginal improvement over the lower threshold. Intuitively, higher region of influence induced by higher threshold is performing better regularization. 
Hence, we decided to use a very high threshold of $250$ pixels, which is almost equivalent to regressing the full distance map and neglecting this hyper-parameter.

{\it Network Weight Distribution: }
We also analyzed the weight distribution of the network before and after distance map regularization, as shown in the Supplementary Materials ({\bf Fig. S5}). We observe the number of non-zero weights increase after the distance map regularization, hence, better utilizing the network capacity. A similar flattening of network weight histogram has been reported for the dropout regularization and Bayesian neural networks \cite{Blundell:2015}, both reducing the overfitting and hence improving generalization. Specifically, the network weights are randomly dropped during dropout, forcing the network to use the remaining weights to identify the patterns in data (spreading the weight histogram), hence creating an ensemble effect with reduced over-fitting and improved generalization. We observe a similar pattern in the weight distribution after the distance map regularization.
% However, this effect is less prominent in the U-Net architecture, as the deconvolution filters are learned rather than using the pooling indices for up-sampling. This gives the U-Net architecture more flexibility to pass information through the skip connections, such that, the regularization imposed at the bottleneck layer has reduced effect.

\section{Conclusion}
% {\it Summary: }
In this work we proposed and implemented a multi-task learning-based regularization method for fully convolutional networks for semantic image segmentation and demonstrated its benefits in the context of cardiac MR image segmentation. To implement the proposed method, we appended a decoder network at the bottleneck layer of existing FCN architectures to perform an auxiliary task of distance map prediction, which is removed after training.
%to reduce inference time.

We automatically learned the weighting of the tasks based on their uncertainty. As the distance map contains robust information regarding the shape, location, and boundary of the object to be segmented, it facilitates the FCN encoder to learn robust global features important for the segmentation task. 

Our experiments verify that introducing the distance map regularization improves the segmentation performance of three FCN architectures for both binary and multi-class segmentation across two publicly available cardiac cine MRI datasets featuring significant patient anatomy and image variability. Specifically, we observed consistent improvement in segmentation performance in the challenging apical and basal slices in response to the soft-constraints imposed by the distance map regularization. We also showed consistent segmentation improvement on all five patient pathology in the ACDC dataset. Furthermore, these improvements were also reflected on the computed clinical indices important for the diagnosis of various heart conditions. Lastly, we demonstrated the proposed regularization significantly improved the generalization ability of the networks on cross-dataset segmentation (transfer learning), without being aware of the new data distribution, with 5\% to 42\% improvement in average Dice coefficient over the baseline FCN architectures.

% \section{Conclusion}
% The conclusion goes here.

% or


% \appendices

% \section{}

\section*{Acknowledgment}
Research reported in this publication was supported by the National Institute of General Medical Sciences of the National Institutes of Health under Award No. R35GM128877 and by the Office of Advanced Cyber infrastructure of the National Science Foundation under Award No. 1808530. Ziv Yaniv's work was supported by the Intramural Research Program of the U.S. National Institutes of Health, National Library of Medicine.

\ifCLASSOPTIONcaptionsoff
  \newpage
\fi



\bibliographystyle{IEEEtran}
\bibliography{bibliography}

\begin{thebibliography}{10}
\providecommand{\url}[1]{#1}
\csname url@samestyle\endcsname
\providecommand{\newblock}{\relax}
\providecommand{\bibinfo}[2]{#2}
\providecommand{\BIBentrySTDinterwordspacing}{\spaceskip=0pt\relax}
\providecommand{\BIBentryALTinterwordstretchfactor}{4}
\providecommand{\BIBentryALTinterwordspacing}{\spaceskip=\fontdimen2\font plus
\BIBentryALTinterwordstretchfactor\fontdimen3\font minus
  \fontdimen4\font\relax}
\providecommand{\BIBforeignlanguage}[2]{{%
\expandafter\ifx\csname l@#1\endcsname\relax
\typeout{** WARNING: IEEEtran.bst: No hyphenation pattern has been}%
\typeout{** loaded for the language `#1'. Using the pattern for}%
\typeout{** the default language instead.}%
\else
\language=\csname l@#1\endcsname
\fi
#2}}
\providecommand{\BIBdecl}{\relax}
\BIBdecl

\bibitem{Peng:2016}
P.~Peng, K.~Lekadir, A.~Gooya, L.~Shao, S.~E. Petersen, and A.~F. Frangi, ``A
  review of heart chamber segmentation for structural and functional analysis
  using cardiac magnetic resonance imaging,'' \emph{Magnetic Resonance
  Materials in Physics, Biology and Medicine}, vol.~29, no.~2, pp. 155--195,
  Apr 2016.

\bibitem{Petitjean2011}
C.~Petitjean and J.-N. Dacher, ``A review of segmentation methods in short axis
  cardiac {MR} images,'' \emph{Medical Image Analysis}, vol.~15, no.~2, pp. 169
  -- 184, 2011.

\bibitem{Long:2015}
J.~Long, E.~Shelhamer, and T.~Darrell, ``Fully convolutional networks for
  semantic segmentation,'' in \emph{The IEEE Conference on Computer Vision and
  Pattern Recognition (CVPR)}, June 2015.

\bibitem{LeCun:1998}
Y.~Lecun, L.~Bottou, Y.~Bengio, and P.~Haffner, ``Gradient-based learning
  applied to document recognition,'' \emph{Proceedings of the IEEE}, vol.~86,
  no.~11, pp. 2278--2324, Nov 1998.

\bibitem{Goodfellow:2016}
I.~Goodfellow, Y.~Bengio, and A.~Courville, \emph{Deep Learning}.\hskip 1em
  plus 0.5em minus 0.4em\relax MIT Press, 2016.

\bibitem{LeCun:2015}
Y.~LeCun, Y.~Bengio, and G.~Hinton, ``Deep learning,'' \emph{Nature}, vol. 521,
  no. 7553, pp. 436--444, 05 2015.

\bibitem{Shen:2017}
D.~Shen, G.~Wu, and H.-I. Suk, ``Deep learning in medical image analysis,''
  \emph{Annual review of biomedical engineering}, vol.~19, pp. 221--248, 06
  2017.

\bibitem{Litjens:2017}
G.~Litjens \emph{et~al.}, ``A survey on deep learning in medical image
  analysis,'' \emph{Medical Image Analysis}, vol.~42, pp. 60 -- 88, 2017.

\bibitem{Tran:2016}
P.~V. Tran, ``A fully convolutional neural network for cardiac segmentation in
  short-axis {MRI},'' \emph{CoRR}, vol. abs/1604.00494, 2016.

\bibitem{Poudel:2016}
R.~P.~K. Poudel, P.~Lamata, and G.~Montana, ``Recurrent fully convolutional
  neural networks for multi-slice {MRI} cardiac segmentation,'' in
  \emph{Reconstruction, Segmentation, and Analysis of Medical Images}.\hskip
  1em plus 0.5em minus 0.4em\relax Cham: Springer International Publishing,
  2017, pp. 83--94.

\bibitem{Avendi:2016}
M.~Avendi, A.~Kheradvar, and H.~Jafarkhani, ``A combined deep-learning and
  deformable-model approach to fully automatic segmentation of the left
  ventricle in cardiac {MRI},'' \emph{Medical Image Analysis}, vol.~30, pp. 108
  -- 119, 2016.

\bibitem{Oktay:2018}
O.~Oktay \emph{et~al.}, ``Anatomically constrained neural networks ({ACNNs}):
  Application to cardiac image enhancement and segmentation,'' \emph{IEEE
  Transactions on Medical Imaging}, vol.~37, no.~2, pp. 384--395, Feb 2018.

\bibitem{Garcia:2017}
A.~Garcia{-}Garcia, S.~Orts{-}Escolano, S.~Oprea, V.~Villena{-}Martinez, and
  J.~G. Rodr{\'{\i}}guez, ``A review on deep learning techniques applied to
  semantic segmentation,'' \emph{CoRR}, vol. abs/1704.06857, 2017.

\bibitem{Krogh:1991}
A.~Krogh and J.~A. Hertz, ``A simple weight decay can improve generalization,''
  ser. NIPS'91.\hskip 1em plus 0.5em minus 0.4em\relax San Francisco, CA, USA:
  Morgan Kaufmann Publishers Inc., 1991, pp. 950--957.

\bibitem{Vincent:2008}
P.~Vincent, H.~Larochelle, Y.~Bengio, and P.-A. Manzagol, ``Extracting and
  composing robust features with denoising autoencoders,'' ser. ICML '08.\hskip
  1em plus 0.5em minus 0.4em\relax New York, NY, USA: ACM, 2008, pp.
  1096--1103.

\bibitem{Srivastava:2014}
N.~Srivastava, G.~Hinton, A.~Krizhevsky, I.~Sutskever, and R.~Salakhutdinov,
  ``Dropout: A simple way to prevent neural networks from overfitting,''
  \emph{Journal of Machine Learning Research}, vol.~15, pp. 1929--1958, 2014.

\bibitem{Ioffe:2015}
S.~Ioffe and C.~Szegedy, ``Batch normalization: Accelerating deep network
  training by reducing internal covariate shift,'' in \emph{Proceedings of the
  32Nd International Conference on International Conference on Machine Learning
  - Volume 37}, ser. ICML'15.\hskip 1em plus 0.5em minus 0.4em\relax JMLR.org,
  2015, pp. 448--456.

\bibitem{Goodfellow:2015}
I.~Goodfellow, J.~Shlens, and C.~Szegedy, ``Explaining and harnessing
  adversarial examples,'' in \emph{International Conference on Learning
  Representations}, 2015.

\bibitem{Caruana:1997}
R.~Caruana, ``Multitask learning,'' \emph{Machine Learning}, vol.~28, no.~1,
  pp. 41--75, Jul 1997.

\bibitem{Bartlett:2003}
P.~L. Bartlett and S.~Mendelson, ``Rademacher and gaussian complexities: Risk
  bounds and structural results,'' \emph{J. Mach. Learn. Res.}, vol.~3, pp.
  463--482, Mar. 2003.

\bibitem{Ruder:2017}
S.~Ruder, ``An overview of multi-task learning in deep neural networks,''
  \emph{arXiv preprint arXiv:1706.05098}, 2017.

\bibitem{Teichmann:2018}
M.~Teichmann, M.~Weber, M.~Zöllner, R.~Cipolla, and R.~Urtasun, ``{MultiNet}:
  Real-time joint semantic reasoning for autonomous driving,'' in \emph{2018
  IEEE Intelligent Vehicles Symposium (IV)}, June 2018, pp. 1013--1020.

\bibitem{Uhrig:2016}
J.~Uhrig, M.~Cordts, U.~Franke, and T.~Brox, ``Pixel-level encoding and depth
  layering for instance-level semantic labeling,'' in \emph{GCPR}, 2016.

\bibitem{Kendall:2017MTL}
A.~Kendall, Y.~Gal, and R.~Cipolla, ``Multi-task learning using uncertainty to
  weigh losses for scene geometry and semantics,'' \emph{CoRR}, vol.
  abs/1705.07115, 2017.

\bibitem{Moeskops:2016}
P.~Moeskops, J.~M. Wolterink, B.~H.~M. van~der Velden, K.~G.~A. Gilhuijs,
  T.~Leiner, M.~A. Viergever, and I.~Isgum, ``Deep learning for multi-task
  medical image segmentation in multiple modalities,'' in \emph{MICCAI}, 2016.

\bibitem{Valindria:2018}
V.~V. Valindria \emph{et~al.}, ``Multi-modal learning from unpaired images:
  Application to multi-organ segmentation in {CT} and {MRI},'' in \emph{2018
  IEEE Winter Conference on Applications of Computer Vision (WACV)}, March
  2018, pp. 547--556.

\bibitem{Xue:2018}
W.~Xue, G.~Brahm, S.~Pandey, S.~Leung, and S.~Li, ``Full left ventricle
  quantification via deep multitask relationships learning,'' \emph{Medical
  Image Analysis}, vol.~43, pp. 54 -- 65, 2018.

\bibitem{Dangi:2019}
S.~Dangi, Z.~Yaniv, and C.~A. Linte, ``Left ventricle segmentation and
  quantification from cardiac cine {MR} images via multi-task learning,'' in
  \emph{STACOM}.\hskip 1em plus 0.5em minus 0.4em\relax Cham: Springer
  International Publishing, 2019, pp. 21--31.

\bibitem{Bai:2017}
M.~Bai and R.~Urtasun, ``Deep watershed transform for instance segmentation,''
  in \emph{2017 IEEE Conference on Computer Vision and Pattern Recognition
  (CVPR)}, July 2017, pp. 2858--2866.

\bibitem{Hayder:2017}
Z.~Hayder, X.~He, and M.~Salzmann, ``Boundary-aware instance segmentation,'' in
  \emph{The IEEE Conference on Computer Vision and Pattern Recognition (CVPR)},
  July 2017.

\bibitem{Bischke:2017}
B.~Bischke, P.~Helber, J.~Folz, D.~Borth, and A.~Dengel, ``Multi-task learning
  for segmentation of building footprints with deep neural networks,''
  \emph{CoRR}, vol. abs/1709.05932, 2017.

\bibitem{Krizhevsky:2012}
A.~Krizhevsky, I.~Sutskever, and G.~E. Hinton, ``{ImageNet} classification with
  deep convolutional neural networks,'' in \emph{Advances in Neural Information
  Processing Systems 25}.\hskip 1em plus 0.5em minus 0.4em\relax Curran
  Associates, Inc., 2012, pp. 1097--1105.

\bibitem{Badrinarayanan:2015}
V.~Badrinarayanan, A.~Kendall, and R.~Cipolla, ``{SegNet}: {A} deep
  convolutional encoder-decoder architecture for image segmentation,''
  \emph{CoRR}, vol. abs/1511.00561, 2015.

\bibitem{Ronneberger:2015}
O.~Ronneberger, P.~Fischer, and T.~Brox, ``{U-Net}: Convolutional networks for
  biomedical image segmentation,'' \emph{CoRR}, vol. abs/1505.04597, 2015.

\bibitem{He:2016CVPR}
K.~He, X.~Zhang, S.~Ren, and J.~Sun, ``Deep residual learning for image
  recognition,'' in \emph{The IEEE Conference on Computer Vision and Pattern
  Recognition (CVPR)}, June 2016.

\bibitem{Borgefors:1986}
G.~Borgefors, ``Distance transformations in digital images,'' \emph{Computer
  Vision, Graphics, and Image Processing}, vol.~34, no.~3, pp. 344 -- 371,
  1986.

\bibitem{Fonseca:2011}
C.~G. Fonseca \emph{et~al.}, ``The cardiac atlas project - an imaging database
  for computational modeling and statistical atlases of the heart,''
  \emph{Bioinformatics}, vol.~27, no.~16, pp. 2288--2295, 2011.

\bibitem{Suinesiaputra:2014}
A.~Suinesiaputra \emph{et~al.}, ``A collaborative resource to build consensus
  for automated left ventricular segmentation of cardiac {MR} images,''
  \emph{Medical Image Analysis}, vol.~18, no.~1, pp. 50 -- 62, 2014.

\bibitem{Bernard:2018}
O.~Bernard \emph{et~al.}, ``Deep learning techniques for automatic {MRI}
  cardiac multi-structures segmentation and diagnosis: Is the problem solved?''
  \emph{IEEE Transactions on Medical Imaging}, vol.~37, no.~11, pp. 2514--2525,
  Nov 2018.

\bibitem{Yaniv:2018}
Z.~Yaniv, B.~C. Lowekamp, H.~J. Johnson, and R.~Beare, ``{SimpleITK}
  image-analysis notebooks: A collaborative environment for education and
  reproducible research,'' \emph{Journal of Digital Imaging}, vol.~31, no.~3,
  pp. 290--303, 2018.

\bibitem{He:2015}
K.~He, X.~Zhang, S.~Ren, and J.~Sun, ``Delving deep into rectifiers: Surpassing
  human-level performance on imagenet classification,'' in \emph{Proceedings of
  the IEEE international conference on computer vision}, 2015, pp. 1026--1034.

\bibitem{Hinton:2012neural}
G.~Hinton, N.~Srivastava, and K.~Swersky, ``Neural networks for machine
  learning lecture 6a overview of mini-batch gradient descent.''

\bibitem{Baumgartner:2018}
C.~F. Baumgartner, L.~M. Koch, M.~Pollefeys, and E.~Konukoglu, ``An exploration
  of {2D} and {3D} deep learning techniques for cardiac {MR} image
  segmentation,'' in \emph{STACOM}.\hskip 1em plus 0.5em minus 0.4em\relax
  Cham: Springer International Publishing, 2018, pp. 111--119.

\bibitem{Khened:2018}
M.~Khened, V.~Alex, and G.~Krishnamurthi, ``Densely connected fully
  convolutional network for short-axis cardiac cine {MR} image segmentation and
  heart diagnosis using random forest,'' in \emph{STACOM}.\hskip 1em plus 0.5em
  minus 0.4em\relax Cham: Springer International Publishing, 2018, pp.
  140--151.

\bibitem{Isensee:2018}
Isensee \emph{et~al.}, ``Automatic cardiac disease assessment on cine-{MRI} via
  time-series segmentation and domain specific features,'' in
  \emph{STACOM}.\hskip 1em plus 0.5em minus 0.4em\relax Cham: Springer
  International Publishing, 2018, pp. 120--129.

\bibitem{Tan:2017}
L.~K. Tan \emph{et~al.}, ``Convolutional neural network regression for
  short-axis left ventricle segmentation in cardiac cine {MR} sequences,''
  \emph{Medical Image Analysis}, vol.~39, pp. 78 -- 86, 2017.

\bibitem{Li:2010}
B.~Li \emph{et~al.}, ``In-line automated tracking for ventricular function with
  magnetic resonance imaging,'' \emph{JACC: Cardiovascular Imaging}, vol.~3,
  no.~8, pp. 860 -- 866, 2010.

\bibitem{Fahmy:2012}
A.~S. Fahmy, A.~O. Al-Agamy, and A.~Khalifa, ``Myocardial segmentation using
  contour-constrained optical flow tracking,'' in \emph{STACOM}.\hskip 1em plus
  0.5em minus 0.4em\relax Springer Berlin Heidelberg, 2012, pp. 120--128.

\bibitem{Jolly:2012}
M.-P. Jolly \emph{et~al.}, ``Automatic segmentation of the myocardium in cine
  {MR} images using deformable registration,'' in \emph{STACOM}.\hskip 1em plus
  0.5em minus 0.4em\relax Springer Berlin Heidelberg, 2012, pp. 98--108.

\bibitem{Margeta:2012}
J.~Margeta, E.~Geremia, A.~Criminisi, and N.~Ayache, ``Layered spatio-temporal
  forests for left ventricle segmentation from {4D} cardiac {MRI} data,'' in
  \emph{STACOM}.\hskip 1em plus 0.5em minus 0.4em\relax Springer Berlin
  Heidelberg, 2012, pp. 109--119.

\bibitem{Peters:2018}
T.~Peters, C.~Linte, Z.~Yaniv, and J.~Williams, ``Mixed and augmented reality
  in medicine.''\hskip 1em plus 0.5em minus 0.4em\relax CRC Press, 2018, ch.
  Chapter 16. Augmented and Virtual Visualization for Image-Guided Cardiac
  Therapeutics, pp. 231--250.

\bibitem{Wang:2018}
``Deep visual domain adaptation: A survey,'' \emph{Neurocomputing}, vol. 312,
  pp. 135 -- 153, 2018.

\bibitem{Blundell:2015}
C.~Blundell, J.~Cornebise, K.~Kavukcuoglu, and D.~Wierstra, ``Weight
  uncertainty in neural networks,'' ser. ICML'15.\hskip 1em plus 0.5em minus
  0.4em\relax JMLR.org, 2015, pp. 1613--1622.

\end{thebibliography}

\newpage
\renewcommand{\thefigure}{S\arabic{figure}}
\setcounter{figure}{0}
\setcounter{page}{1}
\pagenumbering{roman}

% \subsection{Segmentation Results}
\begin{figure*}[htb]
  \begin{center}
    \begin{tabular}{c}
          \includegraphics[width=0.55\textwidth]{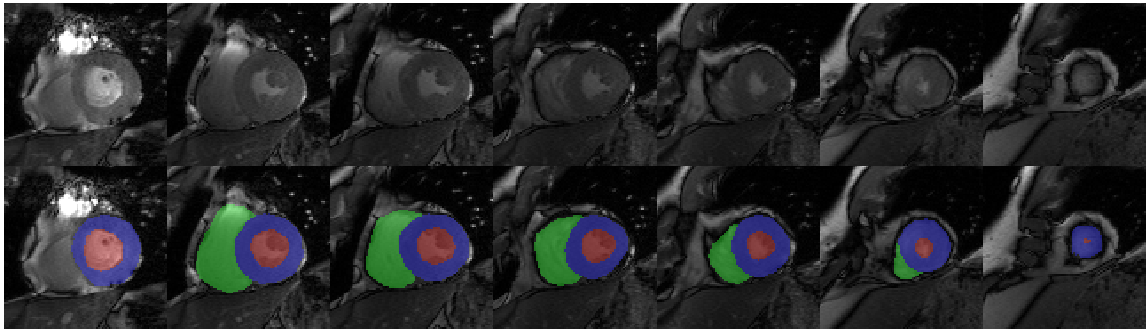}\\
          \vspace{-6mm}\\
          {\footnotesize (a) Input volume (top row) with overlaid ground-truth segmentation (bottom row).} \\
          \vspace{-3mm}\\
          \includegraphics[width=0.55\textwidth]{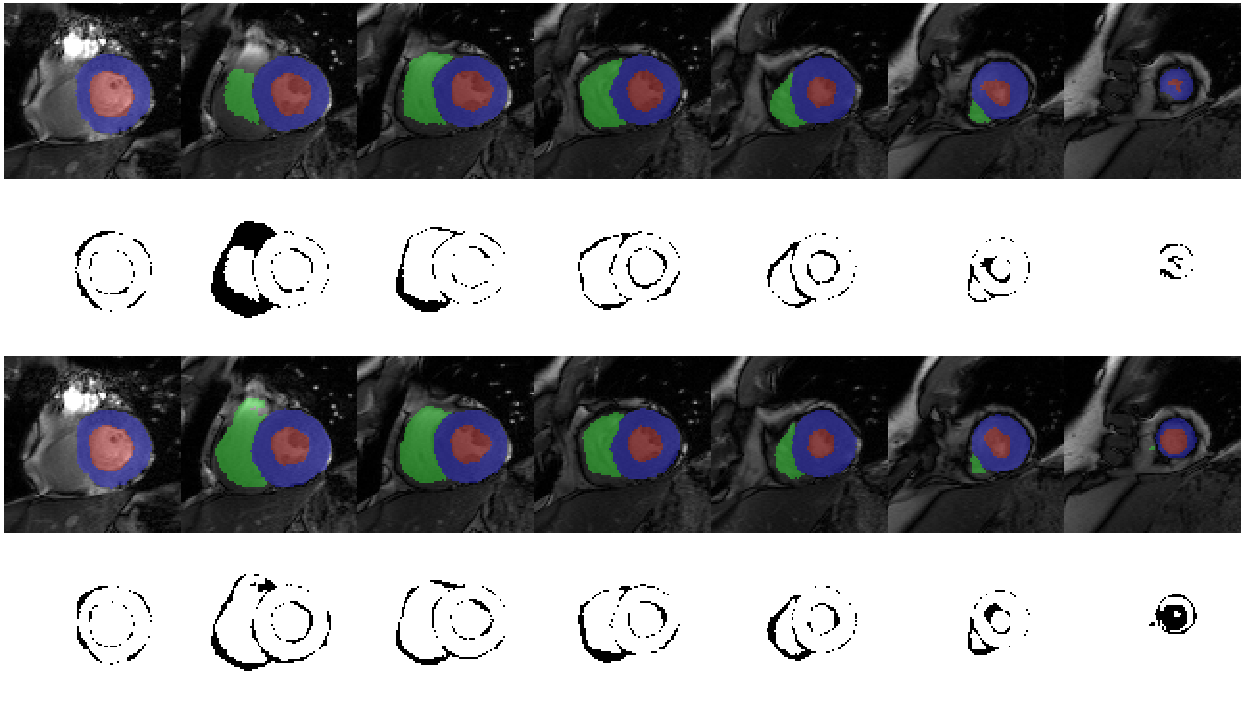}\\
          \vspace{-6mm}\\
          {\footnotesize (b) Segmentation results for SegNet (top two rows) and DMR-SegNet (bottom two rows).} \\
          \vspace{-3mm}\\
          \includegraphics[width=0.55\textwidth]{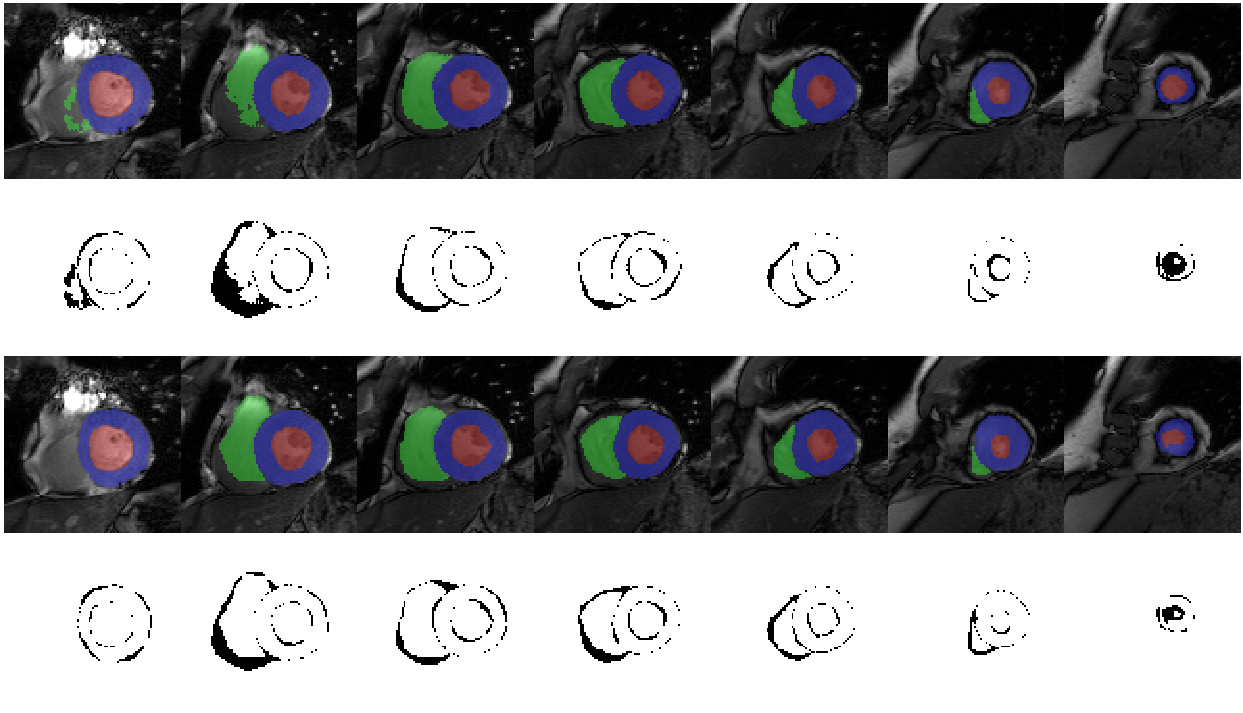}\\
          \vspace{-6mm}\\
          {\footnotesize (c) Segmentation results for USegNet (top two rows) and DMR-USegNet (bottom two rows)} \\
          \vspace{-3mm}\\
          \includegraphics[width=0.55\textwidth]{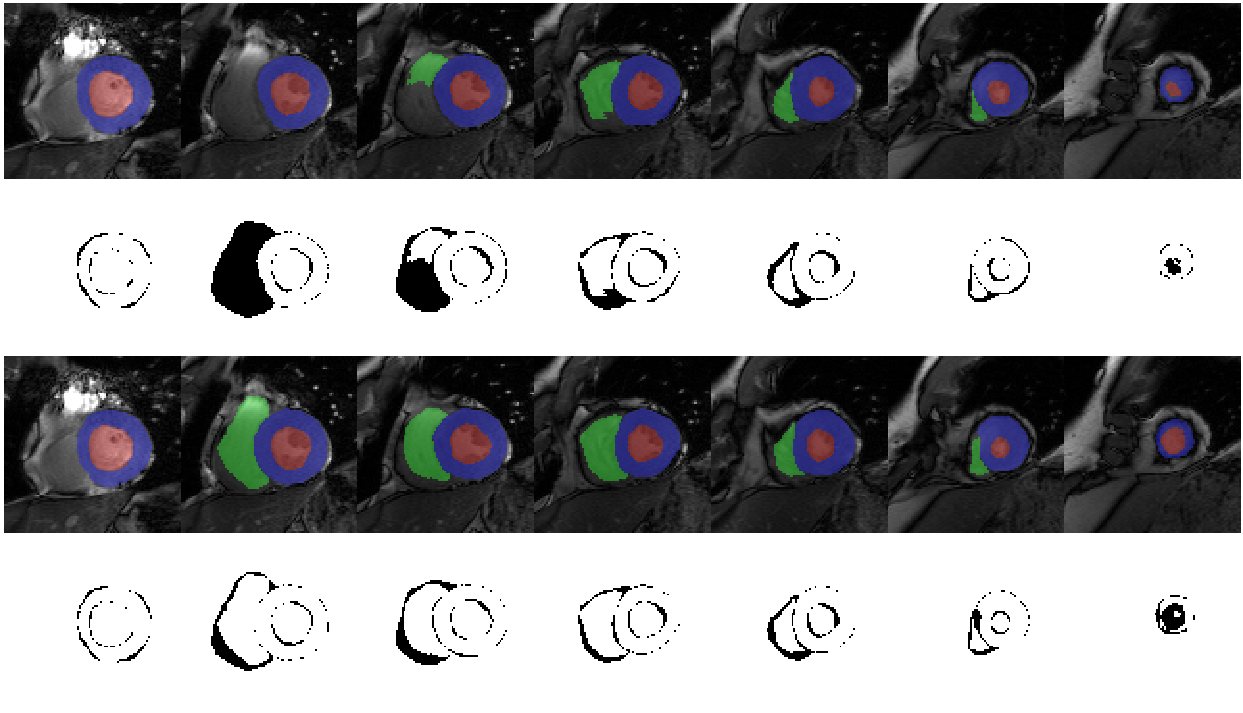}\\
          \vspace{-6mm}\\
          {\footnotesize (d) Segmentation results for UNet (top two rows) and DMR-UNet (bottom two rows)} \\
    \end{tabular}
  \end{center}
  \caption{Ground-truth and automatic segmentation obtained from all trained models for a test patient. In each sub-figure, the segmentation obtained from the baseline and regularized model are overlaid onto the volume and shown in first and third rows, respectively; corresponding disagreement (in black) between the obtained segmentations and the ground-truth is shown in second and fourth rows, respectively.}
  \label{fig:CompareSegmentations}
\end{figure*}

\begin{figure*}[htb]
  \begin{center}
    \begin{tabular}{p{0.865\textwidth}}
          \includegraphics[width=0.75\textwidth]{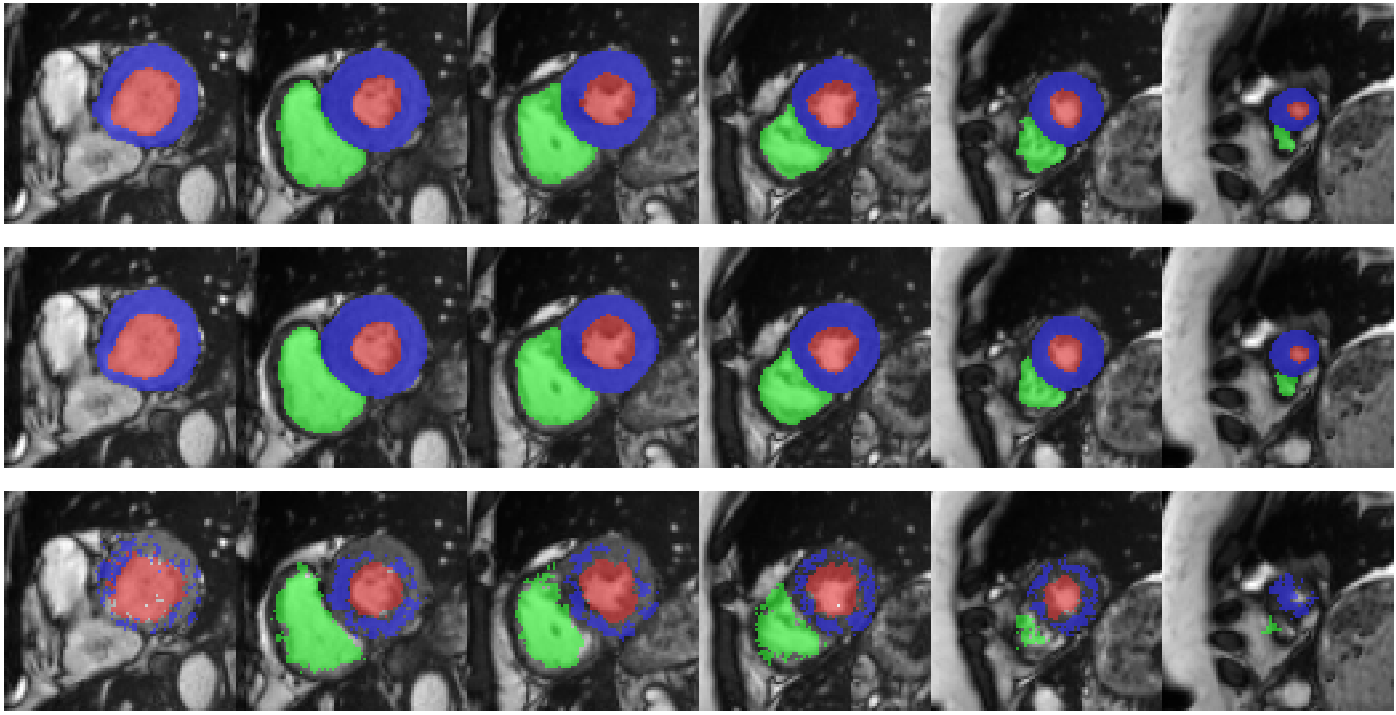}\\
          \vspace{-6mm}\\
          {\footnotesize (a) Input volume with: (top row) ground-truth segmentation overlaid, (middle row) segmentation obtained from the DMR-UNet model, and (bottom row) segmentation obtained after thresholding the predicted distance map at zero levelset.} \\
          \vspace{-3mm}\\
          \includegraphics[width=0.865\textwidth]{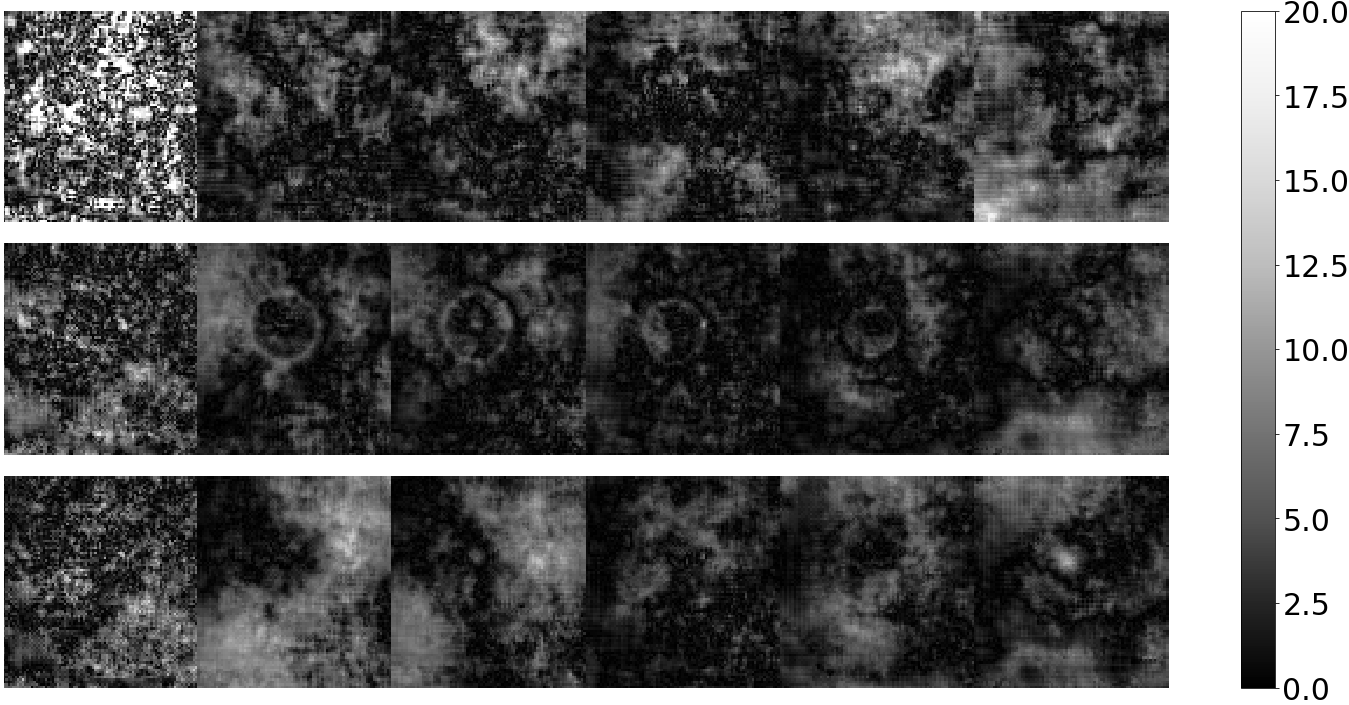}\\
          \vspace{-6mm}\\
          {\footnotesize (b) Absolute difference between the ground-truth and predicted distance maps. First, second, and third row show the error in RV, LV myocardium, and LV bloodpool, respectively.} \\
    \end{tabular}
  \end{center}
  \caption{Visualization of (a) the segmentation obtained by thresholding the predicted distance map and (b) absolute error between the ground-truth and predicted distance maps for all chambers. Shown is only a cropped region around the heart, the error in predicted distance map is higher for the regions farther from the heart.}
  \label{fig:CompareSegmentations}
\end{figure*}

% \FloatBarrier

\begin{figure*}[htb]
  \begin{center}
    \begin{tabular}{cc}
          \includegraphics[width=0.45\textwidth]{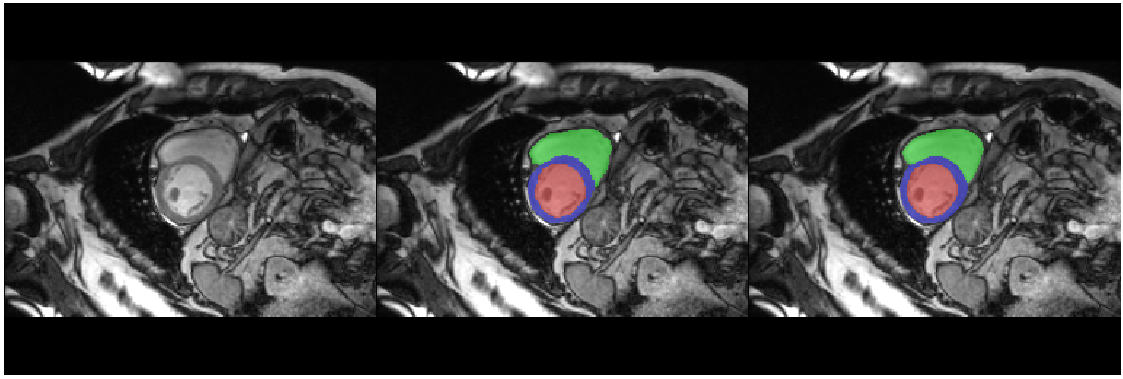} &          
          \includegraphics[width=0.45\textwidth]{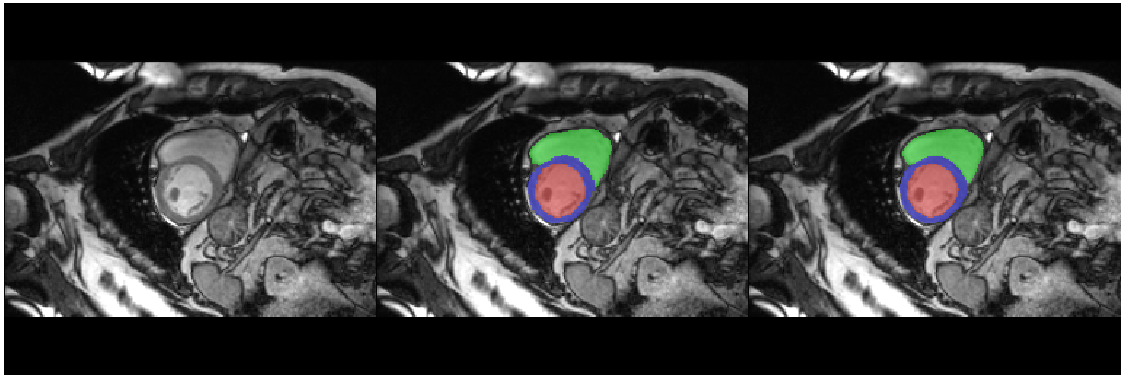} \\
          \multicolumn{2}{c}{(a) From left to right: input image, ground-truth, and automatic segmentation overlay.} \\
          \vspace{0.1mm}\\
          \includegraphics[width=0.45\textwidth]{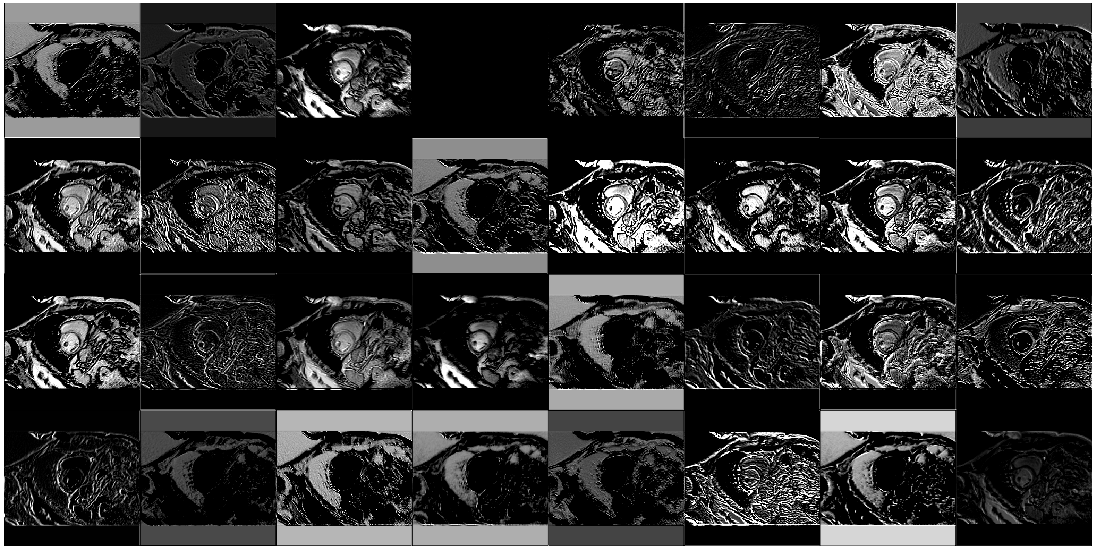} &
          \includegraphics[width=0.45\textwidth]{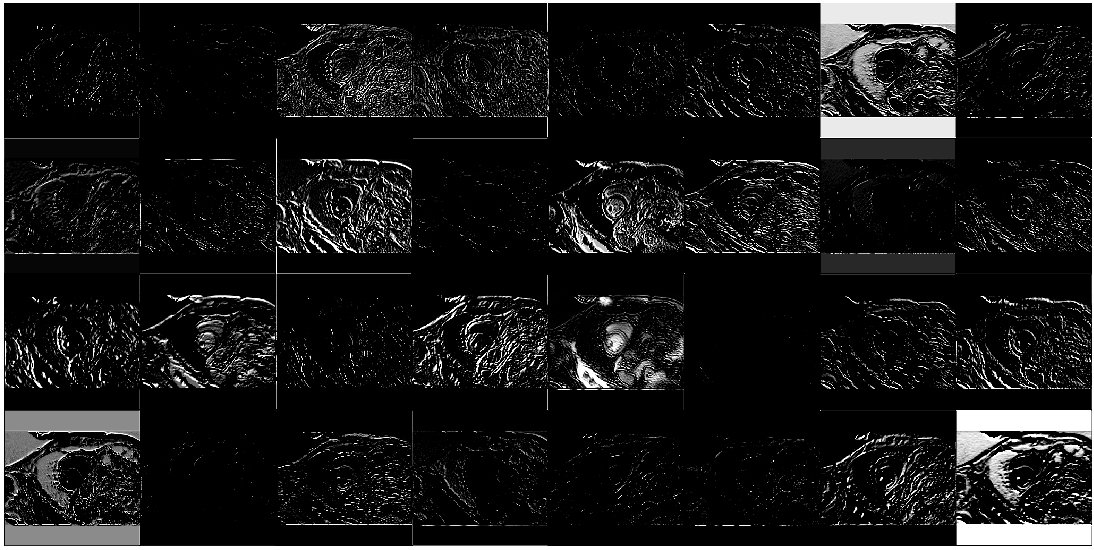} \\
          \multicolumn{2}{c}{(b) 32 feature maps before first max-pooling operation.} \\
          \vspace{0.1mm}\\
          \includegraphics[width=0.45\textwidth]{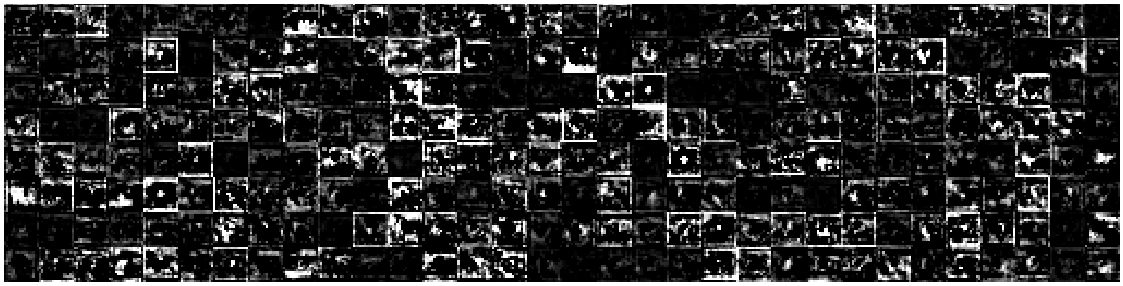} &
          \includegraphics[width=0.45\textwidth]{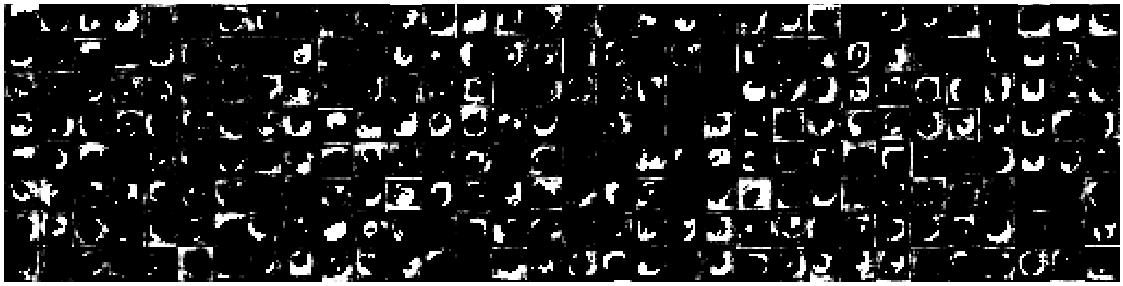} \\
          \multicolumn{2}{c}{(c) 256 feature maps from the bottle-neck layer.} \\
          \vspace{0.1mm}\\
          \includegraphics[width=0.45\textwidth]{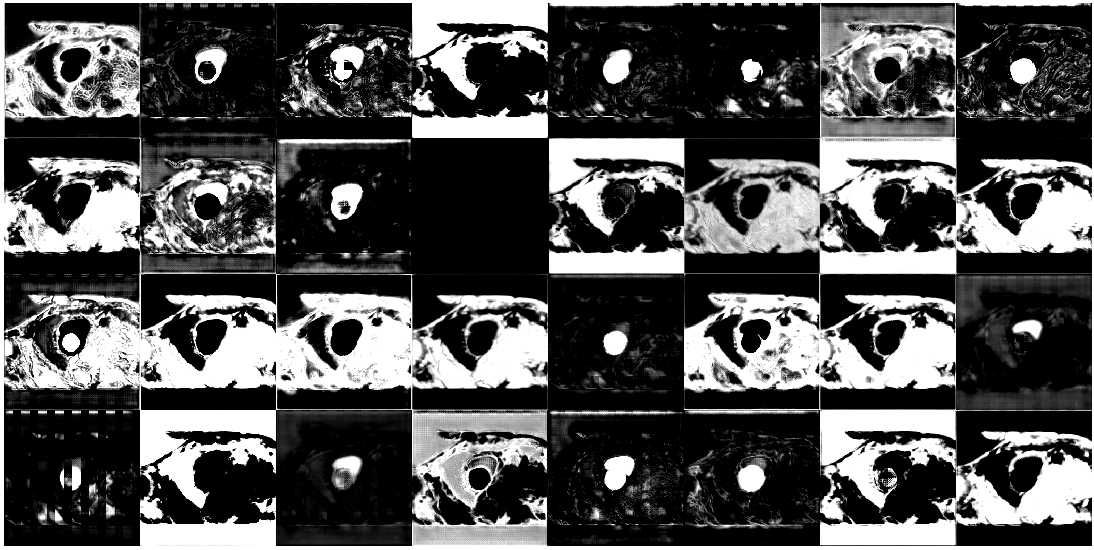} &
          \includegraphics[width=0.45\textwidth]{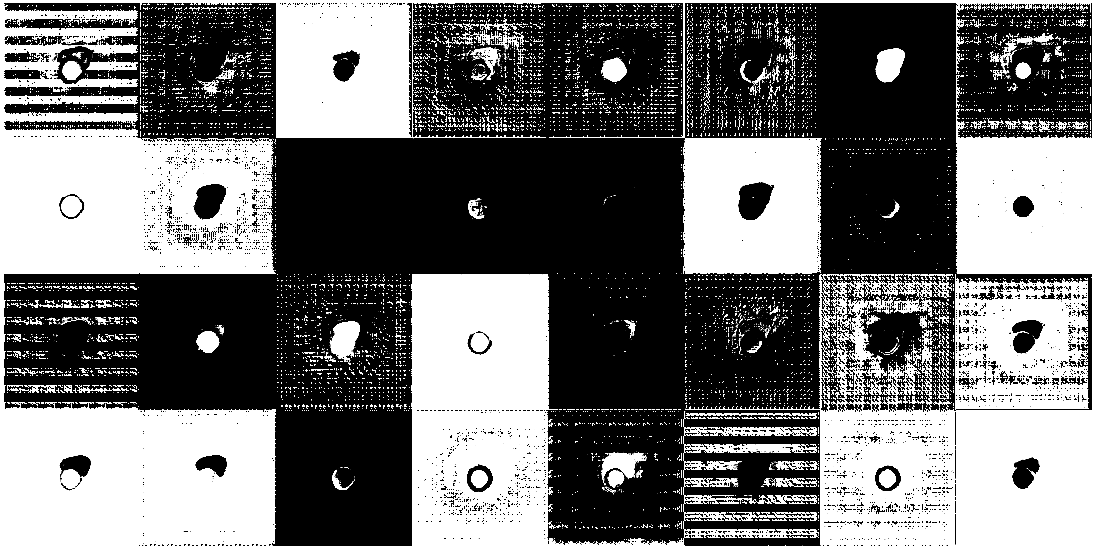} \\
          \multicolumn{2}{c}{(d) 32 feature maps before the final 1$\times$1 convolution.}
    \end{tabular}
  \end{center}
  \caption{Feature maps visualized for the UNet (left column) and DMR-UNet (right column) model. We can observe the UNet model preserves the intensity information and propagates it throughout the network, hence, is more sensitive to the dataset-specific intensity distribution. On the other hand, the DMR-UNet model focuses more on the edges and other discriminative features, producing sparse feature maps, while ignoring dataset-specific intensity distribution. However, the results obtained for intra-dataset segmentation (shown here for ACDC dataset) is similar for both models, whereas, there is a significant improvement in cross-dataset segmentation after distance map regularization.}
  \label{fig:FeatureMapsMT}
\end{figure*}

% \FloatBarrier

% \clearpage
% \subsection{Network Learning Curves}
% Since the cross-entropy loss is harder to interpret, we plot the corresponding dice loss computed during training and testing for both ACDC and LVSC dataset in {\bf Fig. \ref{fig:LearningCurves}a}. We can observe lower difference between the training and test dice loss for the distance map regularized models, demonstrating their ability to prevent overfitting to the training set for the segmentation task. 

\begin{figure*}[tb]
  \begin{center}
    \begin{tabular}{cccc}
          \multicolumn{2}{c}{\includegraphics[width=0.48\textwidth]{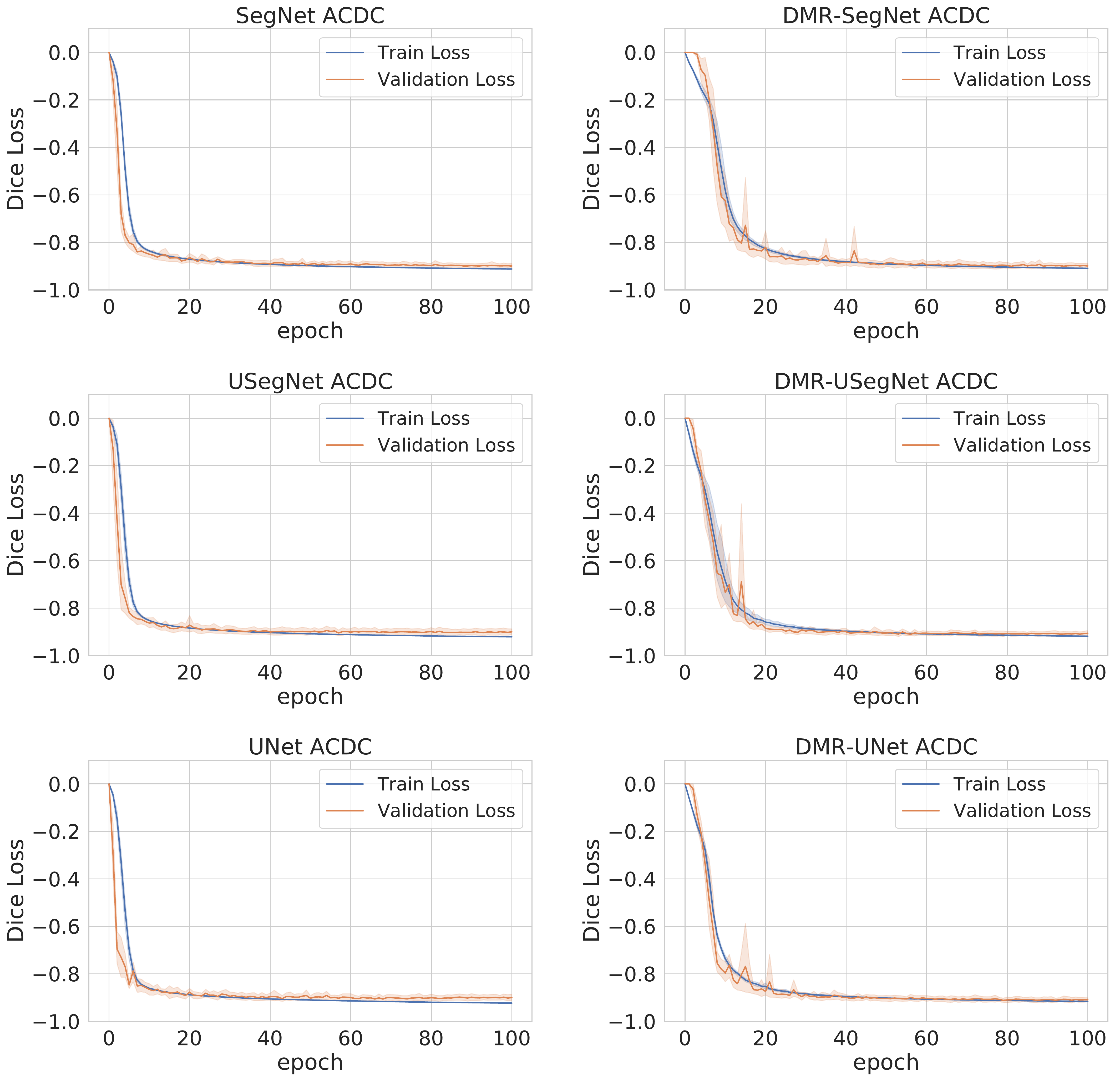}} &          
          \multicolumn{2}{c}{\includegraphics[width=0.48\textwidth]{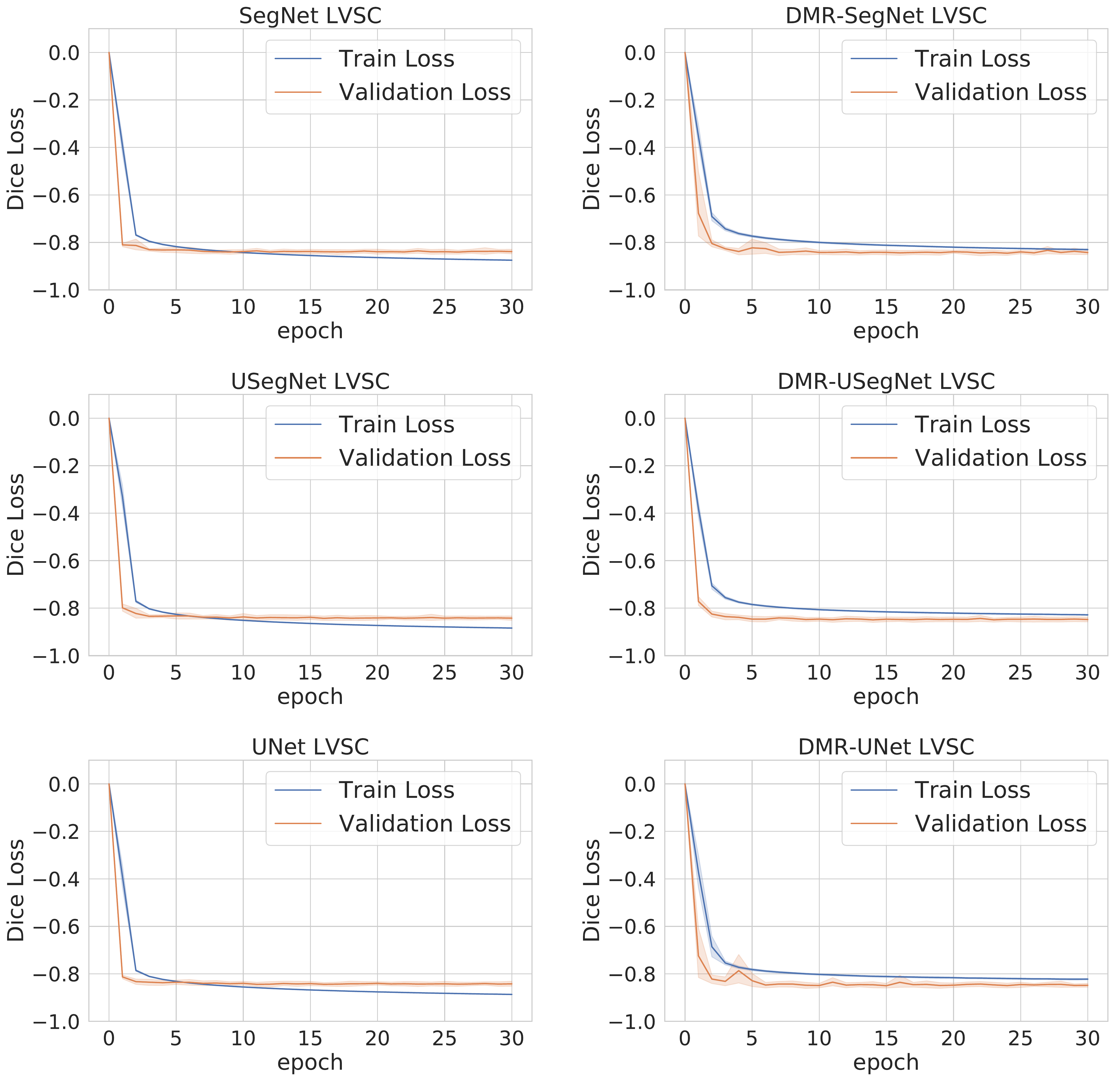}} \\
          \multicolumn{4}{c}{(a) Training and validation Dice loss for segmentation task. ACDC (left two columns) and LVSC (right two columns).} \\
          \vspace{0.1mm}\\
          \includegraphics[width=0.23\textwidth]{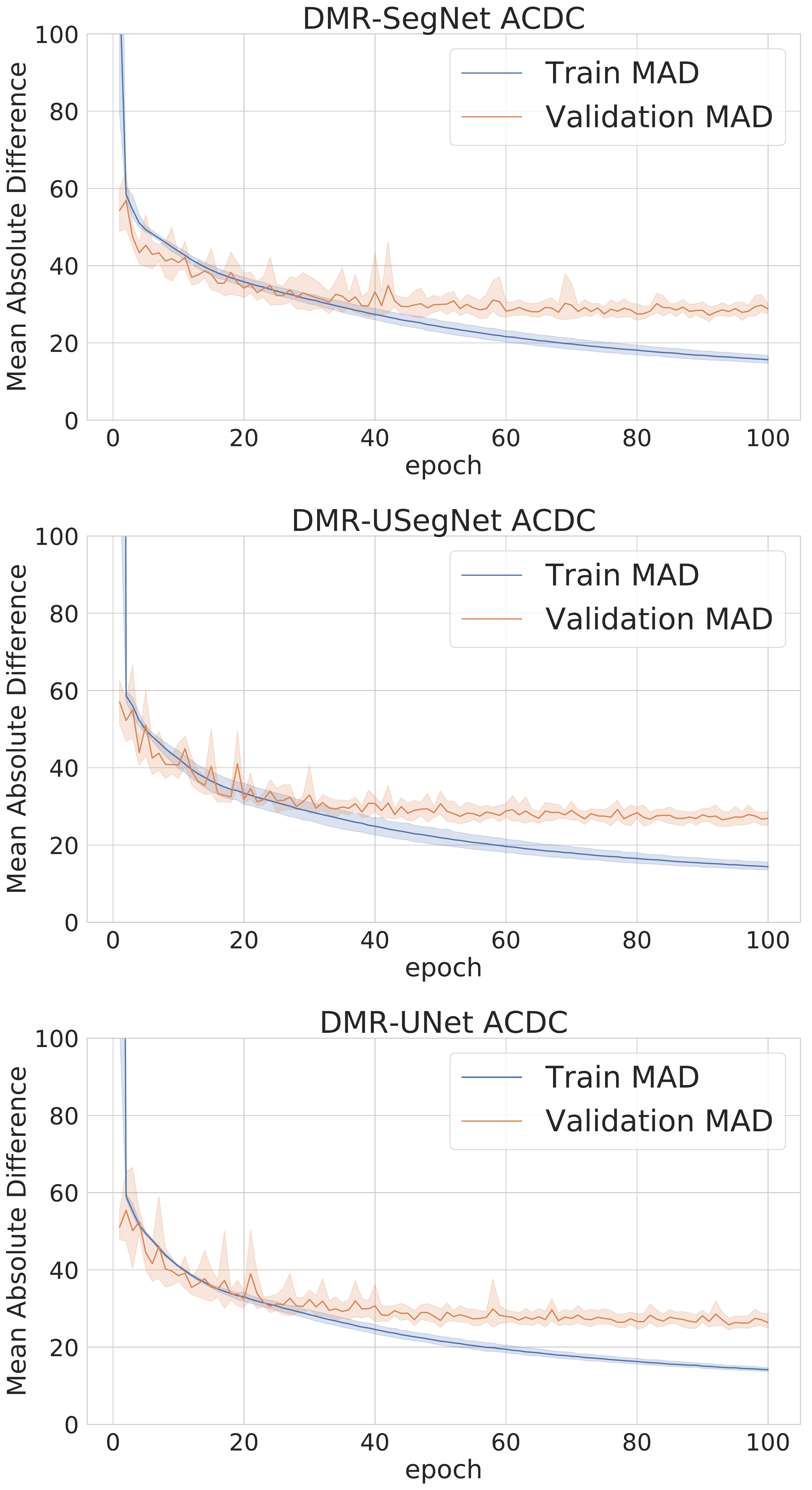} &
          \includegraphics[width=0.23\textwidth]{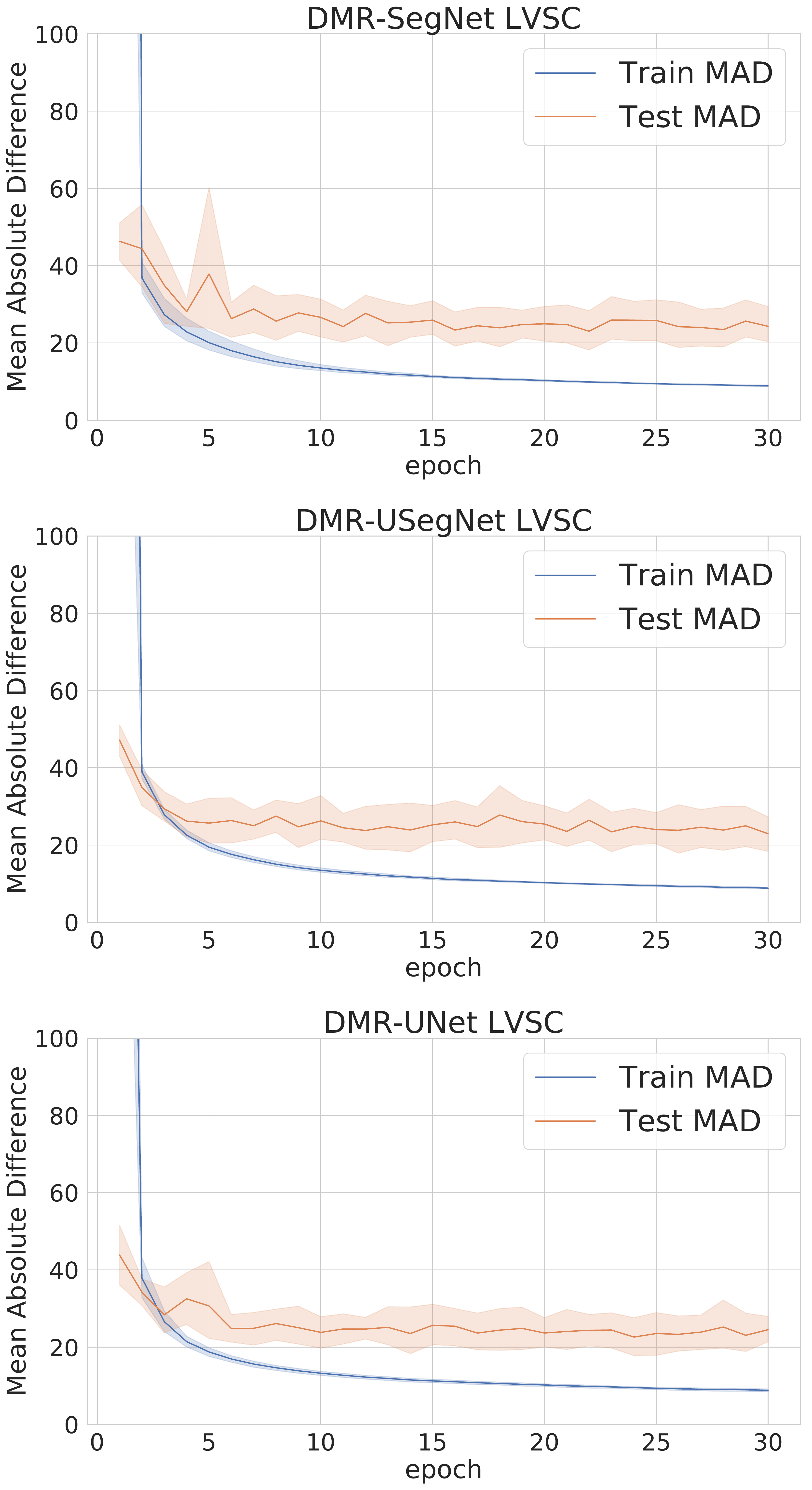} &
          \includegraphics[width=0.23\textwidth]{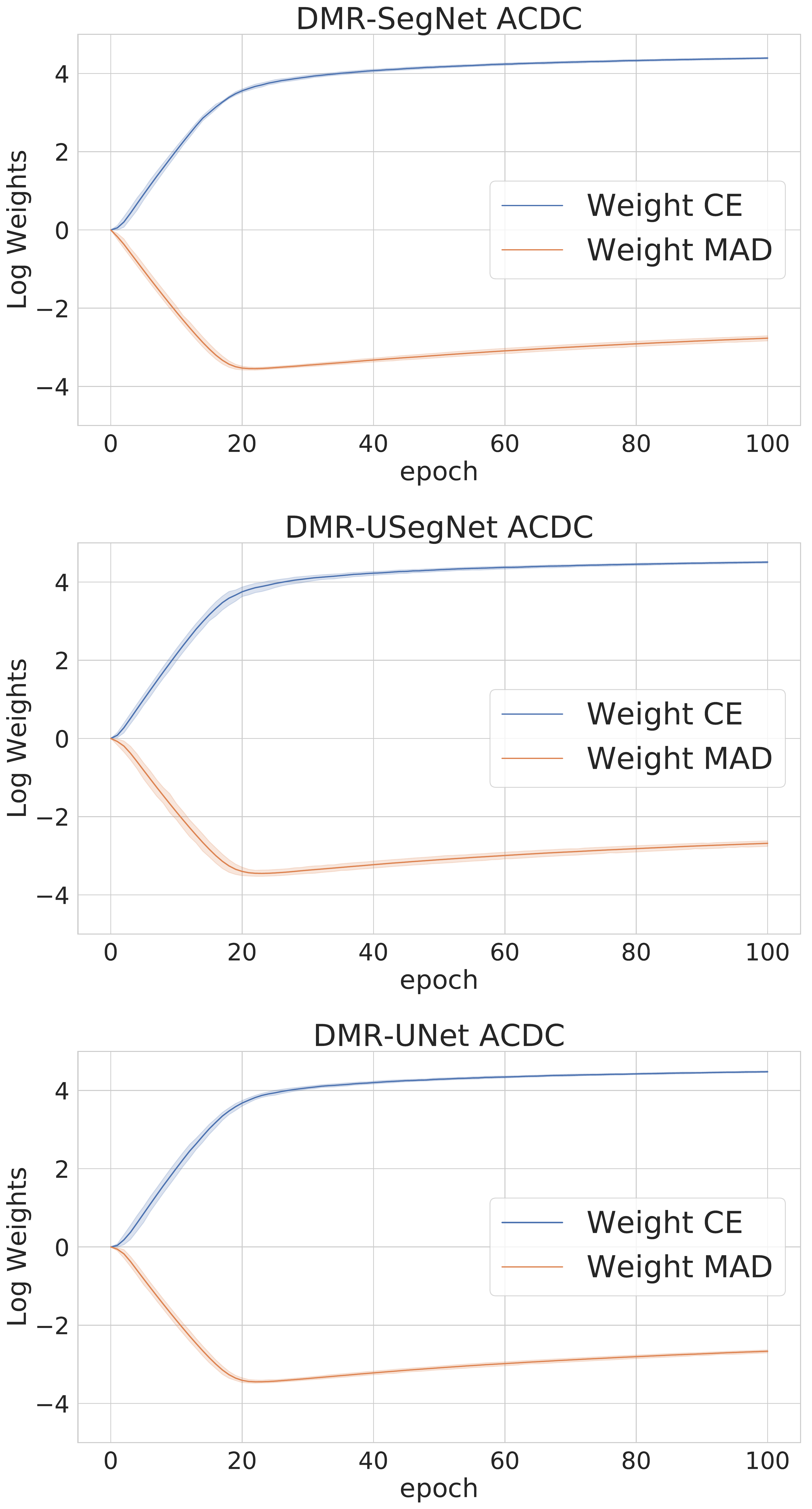} &
          \includegraphics[width=0.23\textwidth]{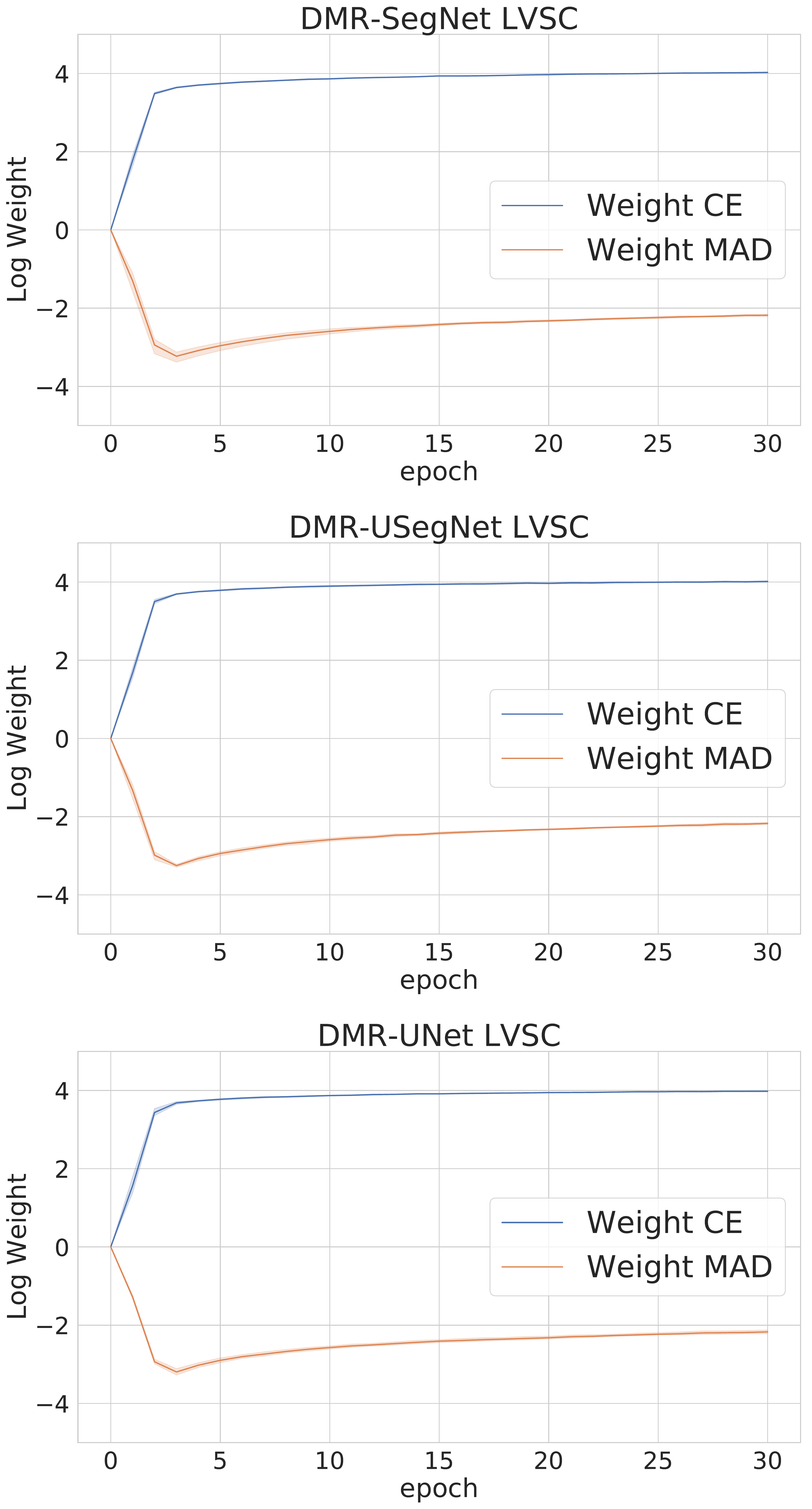} \\
          \multicolumn{2}{p{0.48\textwidth}}{(b) Training and validation mean absolute difference error for distance map regression task. ACDC (left) and LVSC (right).} &
          \multicolumn{2}{p{0.48\textwidth}}{(c) Log Weights learned for cross-entropy and mean absolute difference losses. ACDC (left) and LVSC (right).} \\
    \end{tabular}
  \end{center}
  \caption{Mean and $95\%$ bootstrap confidence interval for training and validation losses (a and b), and the learned weights for cross-entropy and mean absolute difference losses (c), on ACDC and LVSC dataset across five-fold cross-validation. Since the cross-entropy loss is harder to interpret, we plot the corresponding dice loss computed during training and validation. We can observe lower difference between the training and validation dice loss for the distance map regularized models, demonstrating their ability to prevent overfitting.}
  \label{fig:LearningCurves}
\end{figure*}

% \FloatBarrier

% \subsection{Weights Distribution of Trained Networks}

\begin{figure*}[htb]
  \begin{center}
    \begin{tabular}{ccc}
          \includegraphics[width=0.6\textwidth]{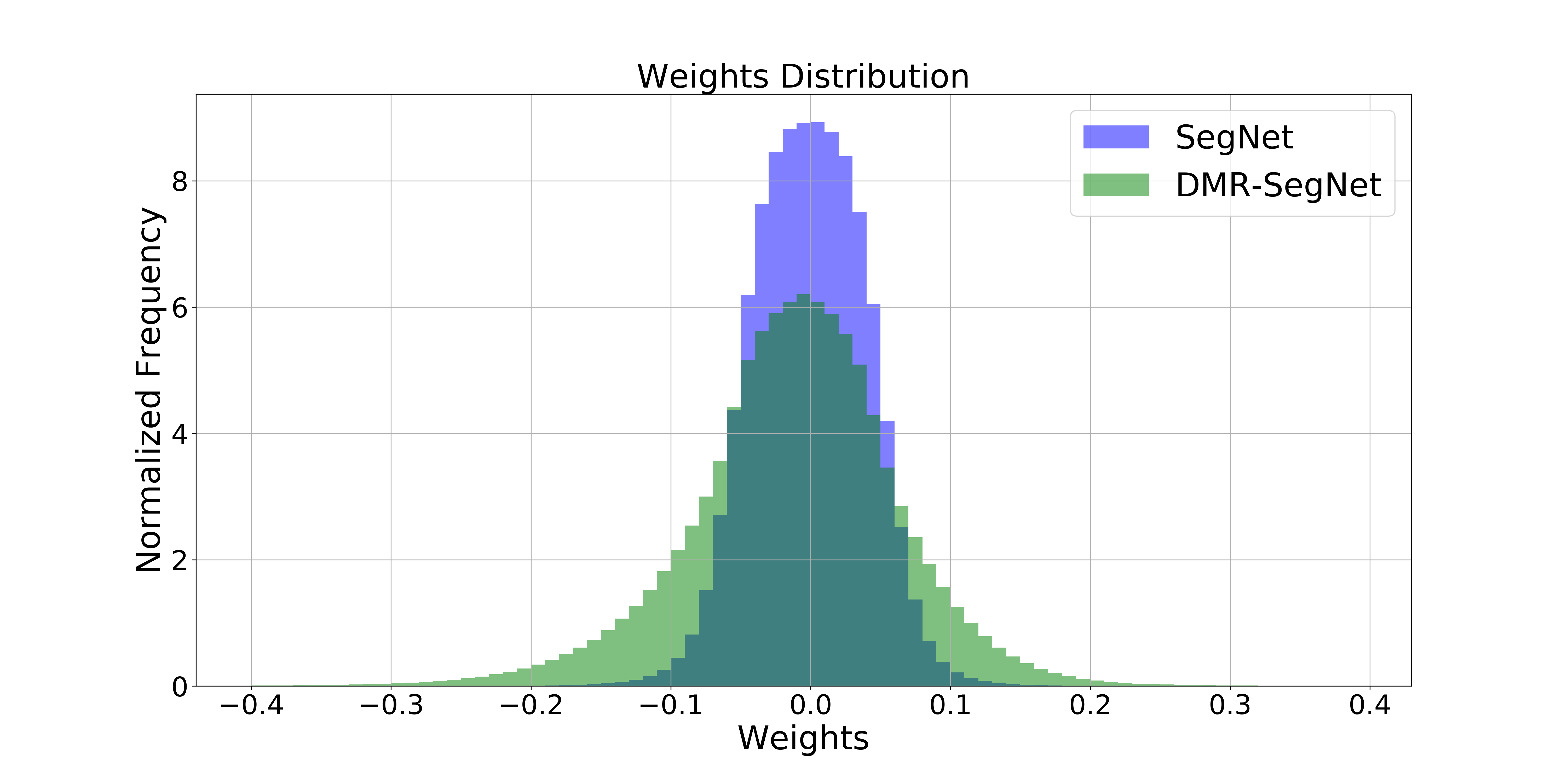}\\
          {\scriptsize(a) Weights distribution for SegNet and DMR-SegNet models.}\\
          \includegraphics[width=0.6\textwidth]{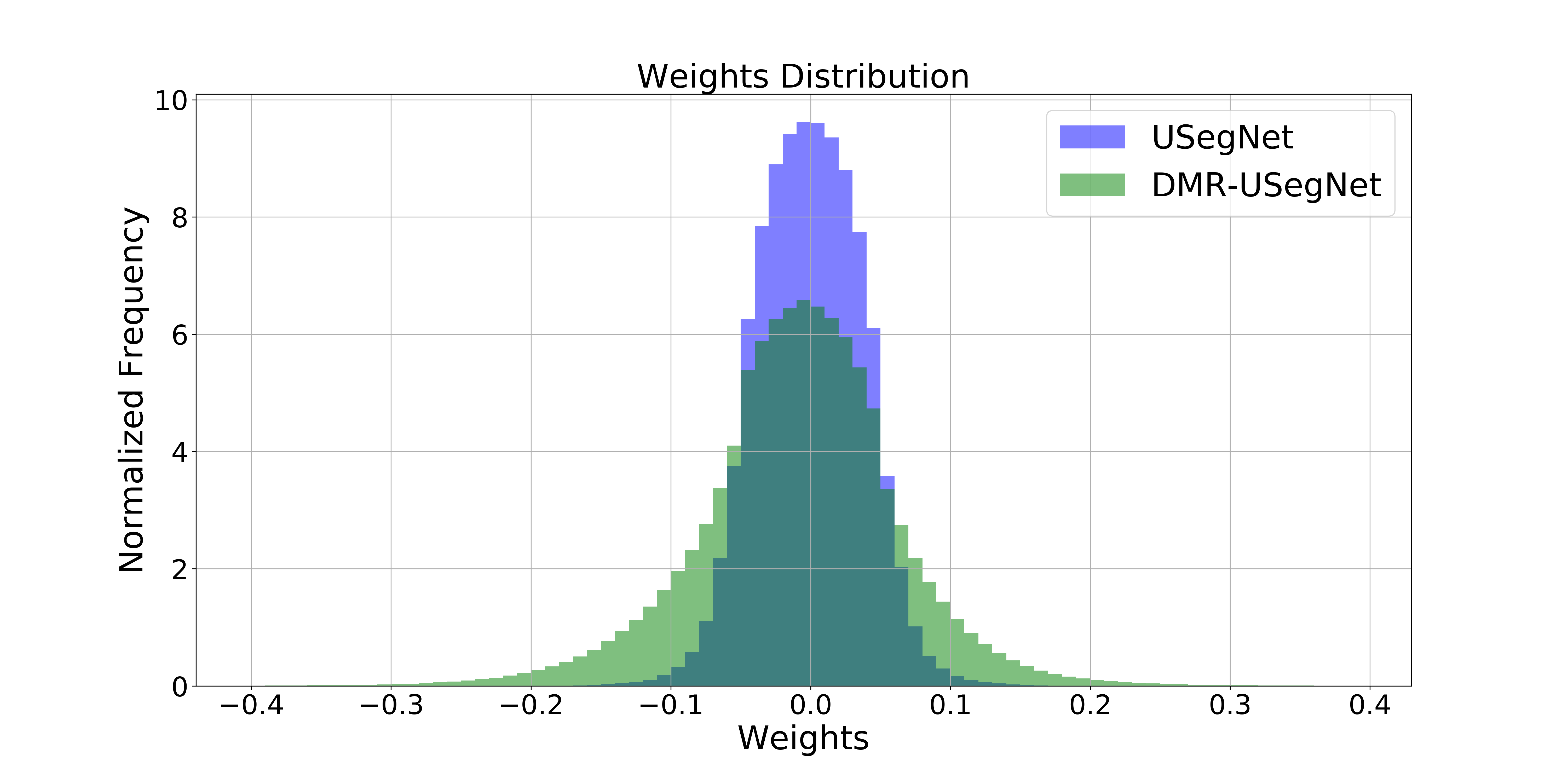}\\
          {\scriptsize(a) Weights distribution for USegNet and DMR-USegNet models.}\\
          \includegraphics[width=0.6\textwidth]{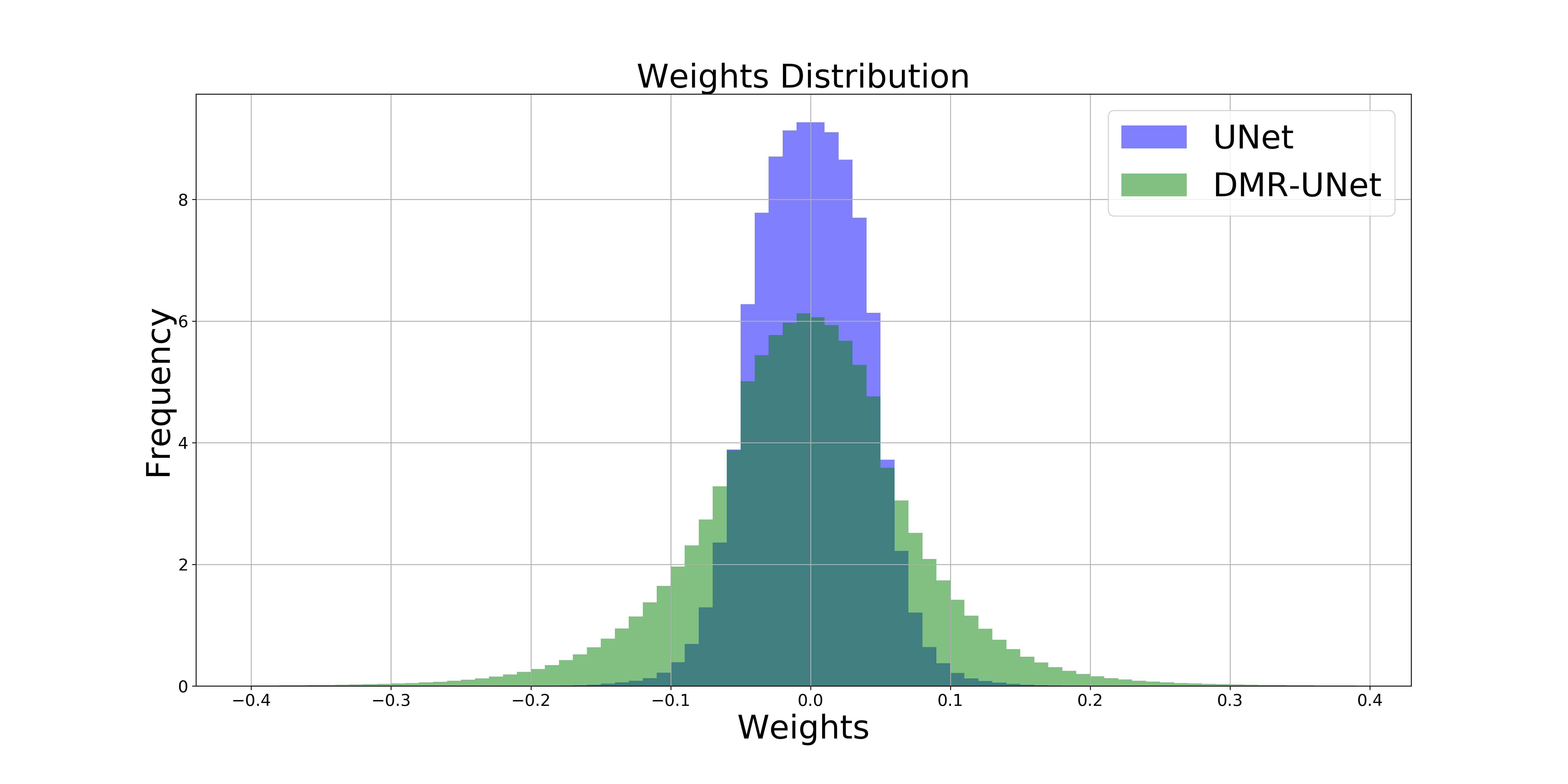}\\
          {\scriptsize(a) Weights distribution for UNet and DMR-UNet models.}\\
    \end{tabular}
  \end{center}
  \caption{Weights distribution before and after distance map regularization for models trained across five-fold cross-validation. We can observe the number of non-zero weights increases after the distance map regularization, hence, better utilizing the network capacity.}
%   The effect of distance map regularization is minimal in UNet architecture (b), as the decoder network does not use the pooling indices, instead, it learns the deconvolution filters. Hence, the network has flexibility to pass information through the skip connections, rather than the bottleneck layer, reducing the effect of the distance map regularization imposed at the bottleneck layer.}
  \label{fig:Weights Distribution}
\end{figure*}

\end{document}